\let\TeXprimitiveyear\year
\documentclass[nolineno]{ieeeaccess}
\let\AccessYearSetter\year
\let\year\TeXprimitiveyear
\usepackage{cite}
\usepackage{amsmath,amssymb,amsfonts}
\usepackage{graphicx}
\usepackage{xpatch}
\usepackage{textcomp,bm}
\usepackage{orcidlink}
\usepackage{booktabs,array,tabularx}
\newcolumntype{D}{>{\centering\arraybackslash}p{.18\columnwidth}}
\newcommand{\gcdatasethead}[1]{{\fontsize{8}{9}\selectfont #1}}
\usepackage{tikz}
\usetikzlibrary{arrows.meta}
\usepackage{etoolbox}
\usepackage{afterpage}
\usepackage{url}
\usepackage{multirow}
\usepackage{booktabs}
\let\year\AccessYearSetter
\definecolor{accessblue}{cmyk}{1,.3,0,.2}
\definecolor{greycolor}{cmyk}{0,0,0,.8}
\definecolor{grey}{cmyk}{0,0,0,.1}
\definecolor{black}{cmyk}{0,0,0,1}
\graphicspath{{figures/}{./}}
\providecommand{\tightlist}{\setlength{\itemsep}{0pt}\setlength{\parskip}{0pt}}
\DeclareTextFontCommand{\texttt}{\ttfamily\fontseries{m}\selectfont}
\newif\ifgcappendixlayout
\gcappendixlayoutfalse

\makeatletter
\pretocmd{\@startcolumn}{%
  \ifgcappendixlayout\else
  \ifx\@dbltoplist\@empty\else
    \global\@topnum\z@\global\@botnum\z@\global\@colnum\z@
  \fi
  \fi
}{}{\PackageError{GraphCompliance}{Could not apply result-float spacing}{Inspect the LaTeX output routine}}
\input{t1pcr.fd}
\DeclareFontShape{T1}{pcr}{n}{n}{<->ssub*pcr/m/n}{}
\AtBeginDocument{\DeclareMathVersion{bold}
\SetSymbolFont{operators}{bold}{T1}{times}{b}{n}
\SetSymbolFont{NewLetters}{bold}{T1}{times}{b}{it}
\SetMathAlphabet{\mathrm}{bold}{T1}{times}{b}{n}
\SetMathAlphabet{\mathit}{bold}{T1}{times}{b}{it}
\SetMathAlphabet{\mathbf}{bold}{T1}{times}{b}{n}
\SetMathAlphabet{\mathtt}{bold}{OT1}{pcr}{b}{n}
\SetSymbolFont{symbols}{bold}{OMS}{cmsy}{b}{n}
\renewcommand\boldmath{\@nomath\boldmath\mathversion{bold}}}
\patchcmd{\@maketitle}{Digital Object Identifier\space\@doi}{}{}{
  \PackageError{GraphCompliance}{Could not suppress sample DOI label}{Inspect official template}}
\let\access@titlepage\ps@titlepage
\let\access@headings\ps@headings
\def\gc@authorfooter{\def\@oddfoot{\hfil{\footerpagefont\thepage}}%
  \def\@evenfoot{{\footerpagefont\thepage}\hfil}}
\def\ps@titlepage{\access@titlepage\gc@authorfooter}
\def\ps@headings{\access@headings\gc@authorfooter}
\makeatother
\hypersetup{pdftitle={GraphCompliance: Structure-Aware Regulatory Compliance through Policy-Context Graph Alignment},
  pdfauthor={Jiseong Chung, Ronny Ko, Won-Yong Shin, Wonchul Yoo, Tae-wan Kim},
  pdfsubject={IEEE Access author-review manuscript}, pdfcreator={LaTeX with official IEEE Access class}}

\begin{document}
\history{}
\doi{}
\title{GraphCompliance: Structure-Aware Regulatory Compliance through Policy--Context Graph Alignment}
\author{\uppercase{Jiseong Chung}\authorrefmark{1}\,
\orcidlink{0009-0005-7575-4119},
\uppercase{Ronny Ko}\authorrefmark{2}\,
\orcidlink{0009-0003-0567-6624},
\uppercase{Won-Yong Shin}\authorrefmark{3}\,
\orcidlink{0000-0002-6533-3469},
\uppercase{Wonchul Yoo}\authorrefmark{1}\,
\orcidlink{0000-0001-6362-2585},
AND
\uppercase{Tae-wan Kim}\authorrefmark{1}\,
\orcidlink{0000-0002-3385-4657}
}
\address[1]{Seoul National University, Seoul, Republic of Korea
(e-mail: jiseong0529@snu.ac.kr; wchyoo@snu.ac.kr; taewan@snu.ac.kr)}
\address[2]{The University of Osaka, Osaka, Japan
(e-mail: ronny@ist.osaka-u.ac.jp)}
\address[3]{Yonsei University, Seoul, Republic of Korea
(e-mail: wy.shin@yonsei.ac.kr)}

\tfootnote{This work was supported by the National Research Foundation of Korea (NRF) grant funded by the Korea government (MSIT) (No. RS-2024-00406127)}

\corresp{Corresponding author: Tae-wan Kim (e-mail: taewan@snu.ac.kr).}
\markboth{GraphCompliance: Policy--Context Graph Alignment}{GraphCompliance: Policy--Context Graph Alignment}
\begin{abstract}
Regulatory compliance is an important requirement in many industrial and administrative settings. Yet compliance assessment remains difficult because regulatory requirements are distributed across structured documents and may depend on hierarchy, definitions, conditions, and cross-references, while the evidence relevant to a particular requirement may be embedded among multiple interacting actors, events, objects, and conditions in the assessed context. We present GraphCompliance, a structure-aware framework that represents regulatory documents as a schema-guided Policy Graph and contextual information as a Context Graph of entity–relation triples. GraphCompliance aligns the two graphs through a Compliance Gate, which identifies relevant regulatory requirements and evaluates them against the corresponding contextual evidence. This design makes regulatory structure explicit while organizing contextual facts into a comparable form, helping preserve structural dependencies and reduce terminology mismatch across different regulatory settings. Evaluation on EU2UK, GDPR-13, and CONTRACT yields Matthews correlation coefficients (MCC) of 0.802, 0.682, and 0.793, respectively. These values are numerically higher than those of the strongest compared methods on all three datasets. Ablation results further show that the key components of GraphCompliance contribute to its overall performance. These results support the utility of structured policy–context alignment for compliance-related prediction across heterogeneous regulatory settings. The source code of GraphCompliance is available at \url{https://github.com/CHUNG-JiSeong/GraphCompliance.git}.
\end{abstract}
\titlepgskip=-21pt
\maketitle
\section{Introduction}

\begin{figure*}[!t]
\centering
\includegraphics[width=\textwidth]{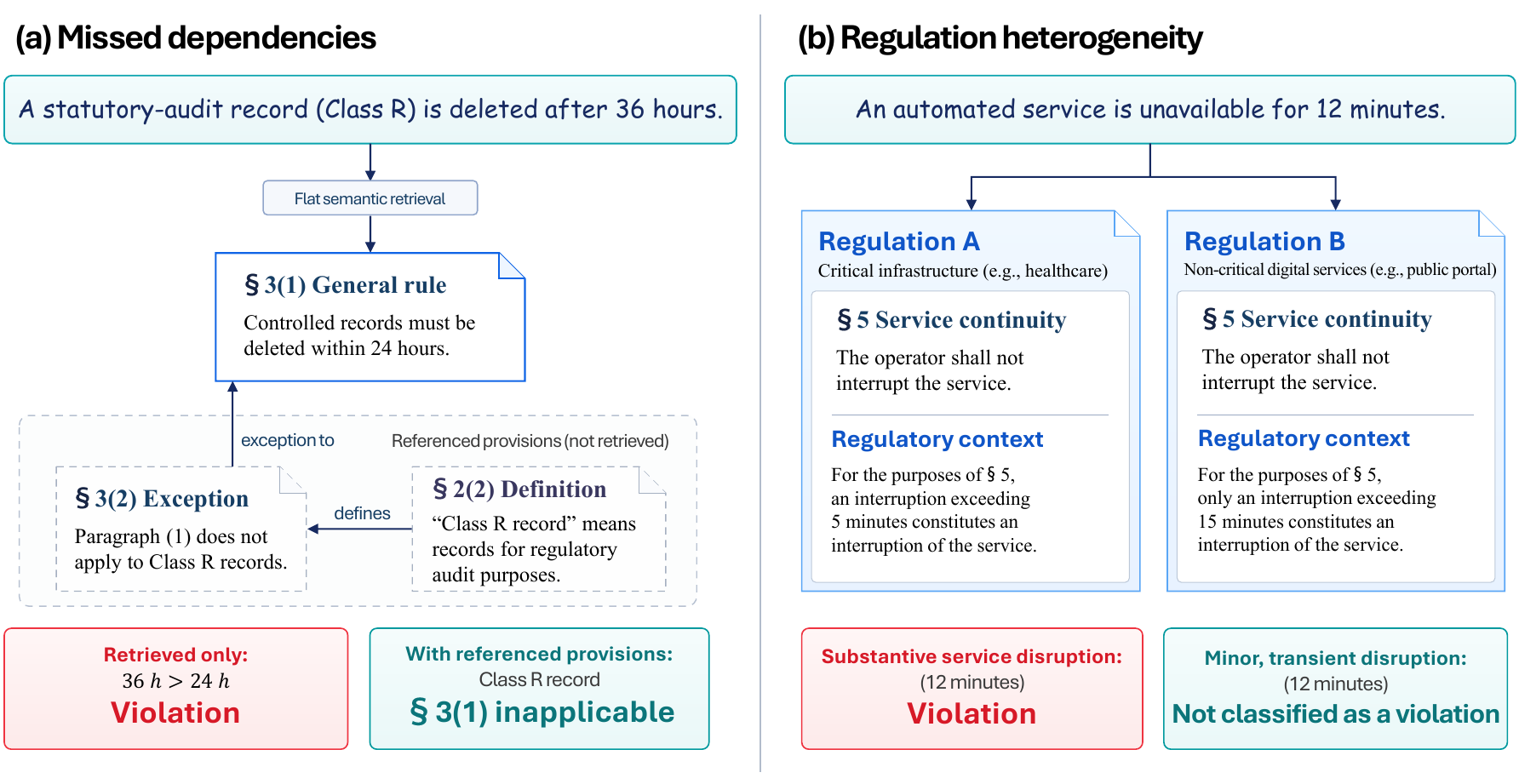}
\caption{Illustrative failure modes in regulatory compliance assessment.
(a) Flat semantic retrieval can apply a general rule outside its scope when linked definitions and exceptions are omitted; incorporating the referenced provisions changes the applicable rule.
(b) Similar regulatory requirements can induce different decision boundaries when their scope and qualifying conditions differ across regulatory settings.}
\label{fig:failure-modes}
\end{figure*}

Regulatory compliance is a persistent requirement in industrial and administrative settings, yet automating its assessment remains challenging. Regulatory documents are organized into chapters, articles, paragraphs, and subordinate provisions. Their interpretation may depend on hierarchy, scope, definitions, conditions, exceptions, and cross-references. The situation being assessed, by contrast, may involve multiple interacting actors, events, objects, and conditions, with evidence relevant to a particular requirement embedded among surrounding contextual details. Compliance analysis therefore requires more than matching two texts: it involves identifying relevant contextual evidence and interpreting its relationship to normative requirements and the circumstances in which they apply \cite{governatori2010norm,athan2015legalruleml}.

Large language models (LLMs) and retrieval-augmented generation (RAG) have been applied to legal and regulatory information access \cite{saeidi2021crosspolicy,sun2025compliance,echenim2025iot}. Many RAG pipelines organize evidence around query-relevant retrieval and response generation \cite{lewis2020rag}, and graph-based variants also support query-focused summarization \cite{edge2024graphrag}. Regulatory assessment, however, presents two distinct challenges. First, semantic similarity alone is insufficient for collecting the complete evidence needed for a regulatory judgment. A provision may inherit its scope from a parent clause, depend on a definition, or be qualified by another requirement, exception, or cross-reference; retrieving a semantically similar passage can therefore omit evidence that governs how the provision should be applied~\cite{koreeda2021contractnli,gokhan2025rirag}. Second, semantically similar regulatory requirements do not necessarily induce the same decision boundary across regulatory settings. Their applicability can vary with the regulatory context, including the institutional setting, scope, governing definitions, conditions, and exceptions of the target regulation. Such variation gives rise to regulatory heterogeneity, where the positive boundary of a policy--context relation may differ across regulations despite similar surface semantics. Thus, both the evidence required for assessment and the boundary between positive and negative policy--context relations must be determined relative to the structure and regulatory context of the target regulation rather than from textual similarity alone.

Figure~\ref{fig:failure-modes} illustrates these two challenges. In Fig.~\ref{fig:failure-modes}(a), retrieving the general rule alone suggests a violation, whereas the linked exception and definition make that rule inapplicable. A retrieval-centered approach can therefore operate on incomplete regulatory evidence when relevant provisions are connected through references but are not retrieved together. Fig.~\ref{fig:failure-modes}(b) illustrates regulatory heterogeneity: similar regulatory requirements can apply differently when their scope and qualifying conditions differ across regulatory settings. The decision boundary between positive and negative policy--context relations must therefore be determined relative to the target regulation rather than inferred from lexical similarity alone.

We formulate this problem as graph alignment between regulatory requirements and contextual events. \textbf{GraphCompliance} represents a regulatory document as a schema-guided \textbf{Policy Graph} that preserves its normative structure and dependencies, while the assessed situation is represented as a \textbf{Context Graph} of entity--relation triples. The two representations are aligned through a \textbf{Compliance Gate}, which identifies relevant regulatory requirements and evaluates them against the corresponding contextual evidence. During this process, structural and reference information in the Policy Graph is retained as supporting policy evidence, while policy-guided hypernym mapping links key Context Graph entities to concepts grounded in the target regulation. Together, these mechanisms help the alignment model account for regulatory dependencies and terminology differences while preserving the original contextual relations. The resulting policy--context representation can be reused across regulatory settings, while the final decision is adapted to the supervision available in each setting.

We evaluate GraphCompliance on three public benchmarks: EU2UK, GDPR-13, and CONTRACT. GraphCompliance achieves MCCs of 0.802, 0.682, and 0.793, respectively. These values are numerically higher than those of the strongest compared
methods on all three datasets. We further conduct ablation and diagnostic analyses of encoder choice, graph construction, candidate retrieval, graph representation, policy-guided hypernym mapping, and cross-setting adaptation to examine how these components contribute to performance and characterize the proposed alignment design.

The contributions are summarized as follows.

\begin{enumerate}
\def\labelenumi{\arabic{enumi}.}
\tightlist
\item
  \textbf{Graph-Alignment Compliance Pipeline.} We propose a pipeline in which structured regulatory requirements guide contextual evidence selection and the resulting policy--context relations are learned for compliance-related assessment.
\item
  \textbf{Structure-Aware Policy Reasoning.} We preserve regulatory structure and reference information in the Policy Graph so that policy--context alignment can account for dependencies beyond an individual provision.
\item
  \textbf{Policy-Guided Hypernym Mapping.} We link key Context Graph entities to concepts grounded in the target policy to reduce terminology mismatch during alignment.
\item
  \textbf{Compliance Gate for Heterogeneous Regulatory Settings.} The Compliance Gate combines a shared applicability encoder with regulation-specific Positive Supervision Filters, allowing a common policy--context representation to accommodate different positive criteria across regulatory settings.
\end{enumerate}
\section{Related Work}

\subsection{Automated Regulatory Compliance}

Automated compliance analysis has been studied through both explicit rule representations and learned semantic models. Formal and ontology-based approaches translate regulatory requirements into machine-processable rules or structured knowledge and verify target processes or artifacts against them. Examples include GDPR verification for Internet of Things environments \cite{barati2020gdpr}, semantic rule checking in construction \cite{guo2021semantic}, and automated analysis of data-processing agreements against GDPR requirements \cite{amaral2023dpa}. Learning-based work addresses complementary aspects of the problem. Liu et al.~combine privacy-policy sentence classification with Article 13 disclosure analysis \cite{liu2021notified}, while Aires et al.~model contractual norm conflicts through semantic representations \cite{aires2018offsets,aires2019contractual}, and NeuralConflict jointly learns conflict identification and semantic relatedness \cite{huang2024neuralconflict}. Saeidi et al.~further formulate cross-policy compliance through questions over policy conditions and an expression tree that combines their answers \cite{saeidi2021crosspolicy}. These studies demonstrate different ways of combining linguistic interpretation with explicit regulatory conditions. Our work differs in using regulatory document structure to organize the policy evidence that is explicitly aligned with contextual evidence before the final decision.

\subsection{Graph- and LLM-Based Compliance Reasoning}

Graphs have been used with language models to organize evidence, retrieve multi-hop information, and select task-relevant subgraphs. GraphRAG constructs entity and community representations for query-focused summarization \cite{edge2024graphrag}, while HippoRAG and G-Retriever exploit graph connectivity or subgraph retrieval for downstream reasoning \cite{gutierrez2024hipporag,he2024gretriever}. Regulatory applications adopt related ideas in task-specific forms. Elluri et al.~combine authority and referencing-document knowledge graphs with neural classification, summarization, and contextual similarity for GDPR policy analysis \cite{elluri2021bert}. PrivComp-KG connects privacy-policy text with regulatory requirements for knowledge-graph-based verification \cite{garza2024privcompkg}, and Echenim and Joshi combine an IoT-specific regulatory graph with LLM-generated SPARQL queries to support privacy-compliance advice \cite{echenim2025iot}. PoliGraph represents data-collection and subsumption relations across privacy-policy text and distinguishes policy-local ontologies from global ontologies~\cite{cui2023poligraph}. In regulatory question answering, RIRAG-2025 separates passage retrieval from answer generation and evaluates generated answers in terms of source entailment, contradiction avoidance, and obligation coverage~\cite{gokhan2025rirag}. The closest empirical comparator to our setting is Sun et al.~\cite{sun2025compliance}, who combine static knowledge graphs, a regulatory eventic graph, business-text representations, graph-neighborhood fusion, and an LLM judge. Their approach already goes beyond flat passage retrieval by incorporating structured regulatory knowledge. Whereas their final decision is produced by an LLM judge over structured regulatory knowledge, GraphCompliance uses regulatory structure to learn policy-unit--context correspondences before the final benchmark-specific decision.

\subsection{Compliance Benchmarks and Comparative Methods}

Public compliance resources capture different policy--text relationships. Chalkidis et al.~introduce EU2UK and UK2EU for document-level correspondence between EU directives and UK legislation \cite{chalkidis2021doc2doc}. Liu et al.~construct a privacy-policy corpus for identifying GDPR Article 13 disclosure categories and supporting policy-level analysis \cite{liu2021notified}. The CONTRACT lineage originates from research by Aires et al.~on conflicts between contractual norms \cite{aires2018offsets,aires2019contractual}. Subsequent work reuses or extends these resources: Zhang et al.~reuse the Article 13 privacy-policy data \cite{zhang2023bertgdpr}, Ichida and Meneguzzi apply natural-language inference to contractual clauses \cite{ichida2021contract}, and Sheng et al.~evaluate a logic-aided compliance checker on EU2UK, UK2EU, and an Aires-derived contractual resource \cite{sheng2026dynamic}. Sun et al.~evaluate EU2UK, GDPR-13, and CONTRACT within one compliance-checking study \cite{sun2025compliance}. Collectively, these benchmarks cover legislative correspondence, regulatory disclosure identification, and contractual norm-conflict detection, providing complementary settings for evaluating compliance-related reasoning.

Across these settings, GraphCompliance explicitly represents regulatory requirements as Compliance Units within a Policy Graph and aligns them with structured contextual evidence in the Context Graph. A common policy--context alignment mechanism is then adapted to the positive criterion of each benchmark setting.
\section{Methodology}

\begin{figure*}[t]
\centering
\includegraphics[width=\textwidth]{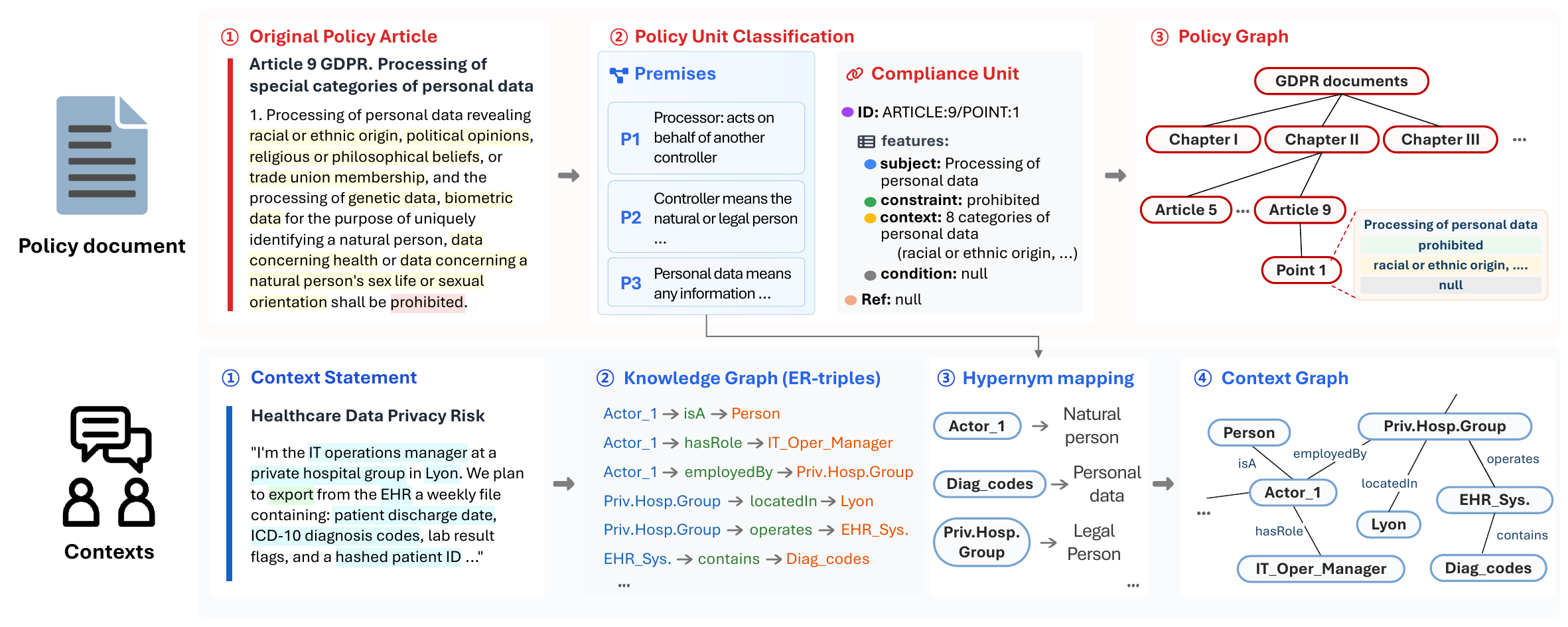}
\caption{Illustrative example of construction of the Policy Graph (top) and Context Graph (bottom). Policy provisions are organized into structural units and Compliance Units with their normative fields and references, while context text is converted into entity--relation triples and enriched with policy-guided hypernyms before alignment.}
\label{fig:graph-construction}
\end{figure*}

GraphCompliance formulates regulatory compliance analysis as alignment between regulatory requirements and contextual events. Its three components are Policy Graph construction, Context Graph construction, and the Compliance Gate. The Policy Graph organizes requirements and structural dependencies, while the Context Graph represents the actors, events, and relationships in the assessed situation. The gate selects relevant pairs and evaluates how regulatory units apply to their contextual evidence. Figure~\ref{fig:graph-construction} illustrates Policy Graph and Context Graph construction.

Let \(P\) denote a target policy document or designated document collection, and let \(x_i\) denote a context. The graph builders produce

\begin{equation}
\label{eq:graphs}
G_P=\mathcal B_P(P),\qquad G_{C,i}=\mathcal B_C(x_i;G_P).
\end{equation}

Conditioning on \(G_P\) denotes policy-guided hypernym mapping, through which Context Graph entities can be associated with higher-level concepts grounded in the target policy. Alignment compares a regulatory unit with a corresponding view of the Context Graph; it does not require exact graph matching or one-to-one node correspondence.

\subsection{Policy Graph Construction}

The Policy Graph is a typed, attributed graph \(G_P=(V_P,E_P,\mathcal A_P)\), where \(V_P\), \(E_P\), and \(\mathcal A_P\) denote nodes, edges, and attributes. A predefined schema distinguishes document structure, premises, Compliance Units, and reference targets. Construction proceeds through hierarchical parsing, policy-unit classification, CU formalization, and reference linking.

Hierarchical parsing retains the levels present in the source, including chapters, articles, paragraphs, and subordinate points. The hierarchy also determines the context used for normative extraction. For example, a list item governed by ``the provider shall notify \ldots{}'' retains the actor and obligation supplied by that parent clause even when they are not repeated in the item. Subordinate provisions are therefore interpreted with the governing text needed to formalize their requirements.

Policy units are classified as premise, actor-CU, or meta-CU. Premises contain definitions and interpretive information that support assessment but are not independently assessed as compliance requirements. Lightweight structural rules over headings identify likely premise sections (e.g., ``Definitions,'' ``Interpretation,'' and ``Terms''). Meta-CUs describe instrument-level applicability, including temporal or territorial scope and entry into force, and are identified through characteristic expressions (e.g., ``entry into force'' and ``this Regulation shall apply''). Remaining normative provisions are primarily treated as actor-CUs, representing obligations, prohibitions, or permissions directed to an actor or regulated entity. The classification stage can optionally use an LLM when greater semantic resolution is preferred; the experimental configuration uses the lightweight rule-based mode for efficiency. CU formalization is performed using an LLM to extract the normative fields from each provision and its governing text.

Actor-CUs and meta-CUs share the representation

\begin{equation}
\label{eq:cu}
u=\langle S_u,\Theta_u,\Pi_u,\kappa_u\rangle,
\end{equation}

where \(S_u\) is the subject, \(\Theta_u\) the condition, \(\Pi_u\) the normative constraint, and \(\kappa_u\) the contextual scope. The constraint retains the substantive requirement rather than only an obligation or prohibition label. Conjunctions, disjunctions, negation, modality, and relevant temporal or quantitative restrictions are retained where they affect interpretation. A field can remain empty when the source provides no corresponding information.

Typed edges connect the units. \texttt{CONTAIN} records hierarchical containment, \texttt{DERIVES} connects a provision to its CUs, and \texttt{REFERS\_TO} represents cross-references. Explicit and relative references are connected to their targets when resolvable, providing explicit and reproducible paths through the regulatory dependencies represented in the Policy Graph. Unresolved references are retained as reference targets.

\subsection{Context Graph Construction}

GraphCompliance constructs the Context Graph as a knowledge graph (KG) from the input context. Nodes represent entities in the described situation, while directed edges represent their relationships. Construction uses GPT-4.1~\cite{openai2025gpt41} for entity--relation extraction in the general GraphRAG setting \cite{edge2024graphrag}. Separating entity identification from relation extraction has been reported to improve entity consistency \cite{mo2025kggen}. We therefore first establish the entity inventory and then extract relations whose endpoints belong to that inventory:

\begin{equation}
\label{eq:extraction}
\begin{aligned}
V_{C,i}&=\operatorname{EntityExtract}(x_i),\\
E_{C,i}&=\operatorname{RelationExtract}(x_i,V_{C,i}),\\
E_{C,i}&\subseteq V_{C,i}\times\mathcal R_C\times V_{C,i},
\end{aligned}
\end{equation}

where \(\mathcal R_C\) is the set of relation labels. Entities may include actors, actions, objects, conditions, and temporal information. Local identifiers bind relation endpoints to the same inventory, retaining the connection between an actor and its surrounding events.

\begin{figure}[t]
\centering
\includegraphics[width=\columnwidth]{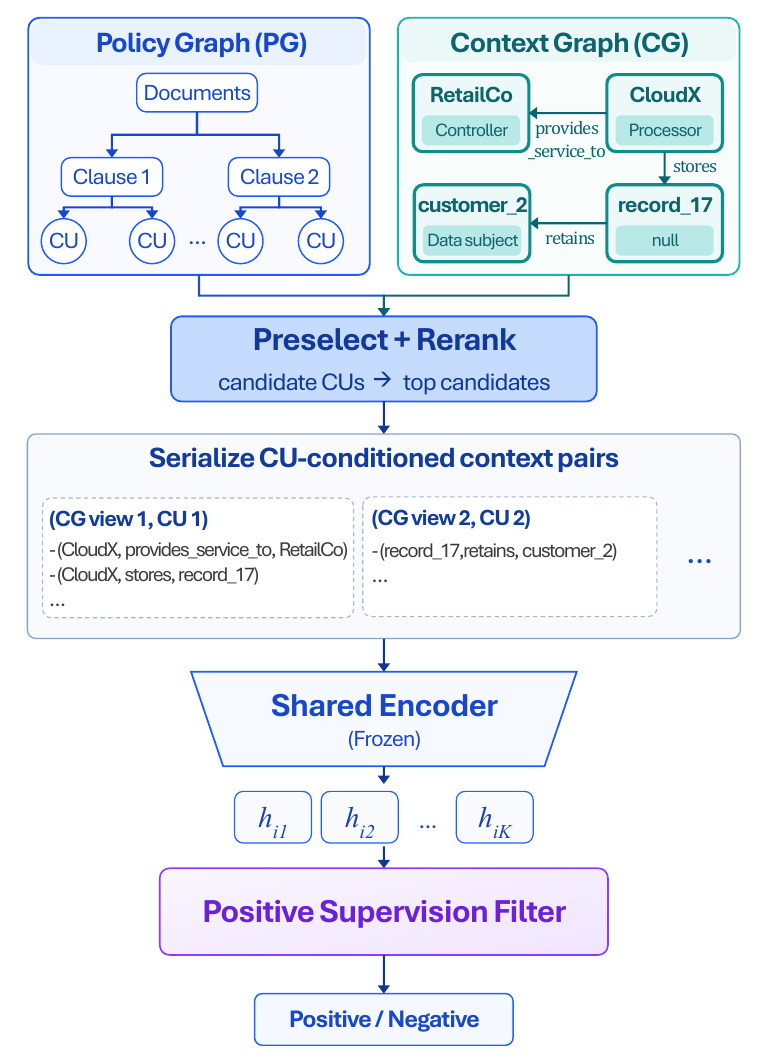}
\caption{Compliance Gate workflow. Candidate CUs are preselected and reranked, paired with CU-conditioned Context Graph views, and encoded by the frozen shared encoder. A regulation-specific Positive Supervision Filter converts each pair representation into an evidence logit, and max-based multiple-instance aggregation produces the sample-level score.}
\label{fig:compliance-gate}
\end{figure}

The initial graph is enriched through policy-guided hypernym mapping. We define this operation as assigning a higher-level concept to a Context Graph entity when that concept is supported by premises retrieved from the target Policy Graph. The mapped concept may represent a regulatory role or another policy-relevant abstraction, and the entity together with its neighboring relations provides the contextual evidence for the assignment. The mapping is auxiliary: it augments the entity representation without replacing the original entity or altering its relations.

Let \(\bar G_{C,i}\) denote the initial Context Graph and \(\mathcal P_P\) the premises identified during Policy Graph construction. For each entity \(e\), relevant premises are retrieved and provided to an LLM together with the entity and its neighboring relations:

\begin{equation}
\label{eq:premise-retrieval}
D_{ie}=\operatorname{RetrievePremises}(e,\bar G_{C,i},\mathcal P_P),
\end{equation}

\begin{equation}
\label{eq:hypernym}
H_P(e)=\operatorname{Map}_{\mathrm{LLM}}(e,\bar G_{C,i},D_{ie}).
\end{equation}

In the final experiments, the policy-guided hypernym mapper \(\operatorname{Map}_{\mathrm{LLM}}\) is implemented using Qwen3-30B~\cite{yang2025qwen3}.

\(H_P(e)\) contains the resulting policy-grounded concepts. Candidate concepts are restricted to terminology supported by the retrieved premises. When no suitable premise or supported concept is available, the mapping remains empty.

For example, consider ``\ldots{} CloudX processes personal data on behalf of RetailCo, the controller \ldots{}''. The definition of processor in GDPR Article 4(8) provides a basis for adding \texttt{Processor} to the \texttt{CloudX} entity \cite{gdpr2016}. This assignment follows the described processing relationship rather than the company name.

Hypernyms are stored as additional attributes without replacing entity names, removing relations, or merging actors that share a regulatory role. The resulting graph preserves the described events while expressing their participants using policy-grounded concepts. This provides an additional semantic bridge between contextual expressions and the terminology of the target regulation before policy--context alignment.

\subsection{Compliance Gate}

The Compliance Gate (Fig.~\ref{fig:compliance-gate}) retrieves candidate CUs, constructs their contextual views, and evaluates the resulting pairs using a pretrained shared applicability encoder and a regulation-specific Positive Supervision Filter. Multiple-instance learning connects the candidate scores to a sample-level decision.

\textbf{Candidate retrieval.} Let $\mathcal{U}_P$ denote the actor-CUs and
meta-CUs in the target Policy Graph. Semantic preselection
followed by reranking produces the top-$K$ candidate set

\begin{equation}
  R_i = \mathrm{Retrieve}_K(G_{C,i}, \mathcal{U}_P), \qquad
  R_i \subseteq \mathcal{U}_P,\ |R_i| \le K,
\end{equation}
where $K$ is the retrieval budget, i.e., the number of reranked
candidate CUs supplied to the Compliance Gate.

Retrieval narrows the comparison space but does not determine compliance: a relevant CU can still concern a different actor, condition, or requirement. The CU representation retains its subject, condition, constraint, and contextual scope. When available, reference information preserved during Policy Graph construction is also included in the serialized CU representation, allowing the alignment model to consider the candidate requirement together with its recorded references. After semantic reranking, each retained candidate is deterministically expanded along Policy Graph reference and hierarchy relations: the referenced or structurally related governing and subordinate CUs needed to preserve its local regulatory context are serialized together with the candidate within the same pair input, so that expansion enriches each
pair without increasing the number of candidate pairs beyond $K$.

\textbf{Global and Local context views.} Global mode retains the complete Context Graph. Local mode selects a neighborhood around actors and events relevant to a candidate CU:

\begin{equation}
\label{eq:context-view}
G_{C,i}^{(u)}=
\begin{cases}
G_{C,i}, & \text{Global mode},\\[2pt]
G_{C,i}[\mathcal N_n(Q_{iu})], & \text{Local mode}.
\end{cases}
\end{equation}

\(Q_{iu}\) denotes the selected actor and event seeds, and \(\mathcal N_n(Q_{iu})\) their neighborhood within \(n\) hops. Selected edges preserve their endpoints and directions. Local mode provides a focused view when the complete graph exceeds the input budget or when interconnected events make the relevant actor's evidence difficult to distinguish. The experimental configuration uses \(n=3\).

\textbf{Shared applicability encoder.} We pretrain a shared applicability encoder on the EU2UK training set using document-level EU--UK correspondence supervision. Because the available supervision is defined at the document-pair level rather than for individual Compliance Units, retrieved CU--Context Graph pairs are treated as a multiple-instance bag; individual CUs in a positive document are not independently labeled as positive. The resulting encoder parameters, denoted by \(\theta^\ast\), are then frozen before regulation-specific adaptation. For each retrieved pair,

\begin{equation}
\label{eq:encoder}
\begin{aligned}
h_{iu}=F_{\theta^\ast}\!\bigl(&\operatorname{ser}_C(G_{C,i}^{(u)}),\\
                           &\operatorname{ser}_P(u)\bigr),
\end{aligned}
\end{equation}

where serialization retains CU fields, available reference information, entity identifiers, and directed relationships. The representation \(h_{iu}\) captures regulation--context applicability: how the CU's subject, constraint, and conditions correspond to actors, events, and circumstances in the context. This includes whether the actor has the required role and whether the contextual event and conditions correspond to those regulated by the CU. The shared representation captures this applicability relation, while benchmark-specific positive criteria are handled by the subsequent decision layer.

\textbf{Regulation-specific Positive Supervision Filter and MIL.} A regulation-specific Positive Supervision Filter converts the shared representation into an evidence logit:

\begin{equation}
\label{eq:head}
z_{iu}^{(d)}=g_{\psi_d}(h_{iu}).
\end{equation}

Here, \(d\) denotes the dataset and \(\psi_d\) the trainable filter parameters. The filter learns how the encoded applicability relation supports the positive criterion of that dataset. In our implementation, \(g_{\psi_d}\) is a linear task-specific head, and only \(\psi_d\) is optimized during target-specific adaptation.

All retrieved pairs associated with a context form a bag:

\begin{equation}
\label{eq:bag}
B_i=\{(G_{C,i}^{(u)},u):u\in R_i\}.
\end{equation}

The bag formulation addresses the negative supervision that can accumulate when candidate pairs are trained independently. A positive sample may retrieve one CU that explains its label and several that do not. Labeling all remaining pairs as independent negatives can overweight negative training signals and can introduce false-negative labels. Bag-level supervision instead assigns the objective at the sample level without requiring every retrieved CU to be positive, following the multiple-instance setting \cite{dietterich1997mil,ilse2018mil}.

For nonempty candidate bags, hard-max aggregation gives the bag-level score

\begin{equation}
  s_i^{(d)} = \max_{u \in R_i} z_{iu}^{(d)}.
  \label{eq:bag_score}
\end{equation}

Given the sample label \(y_i\in\{0,1\}\), the filter is trained using binary cross-entropy,

\begin{equation}
\label{eq:loss}
\begin{aligned}
\mathcal L_i^{(d)}={}&-y_i\log\sigma(s_i^{(d)})\\
 &-(1-y_i)\log\bigl(1-\sigma(s_i^{(d)})\bigr),
\end{aligned}
\end{equation}

where \(\sigma\) is the sigmoid function. A positive bag can be supported by a high-scoring candidate without assigning positive labels to all CUs. In a negative bag, a high candidate score increases the loss through the maximum. This differs from imposing an independent negative loss on every pair.

At inference, both the shared encoder and the adapted filter
are fixed, and the bag-level score $s_i^{(d)}$ is computed with
the same max operator as in~\eqref{eq:bag_score}. The prediction is

\begin{equation}
  \hat{y}_i = \delta_d\bigl(s_i^{(d)}\bigr).
\end{equation}

For each dataset \(d\), the decision rule is \(\delta_d(s)=\mathbf{1}[s\ge\tau_d]\). The threshold \(\tau_d\) is selected on the development set to maximize MCC and is then fixed for final evaluation. Overall, the Compliance Gate separates a reusable representation of policy--context applicability from regulation-specific decision adaptation, allowing the same alignment mechanism to support different positive criteria across regulatory settings.
\section{Experimental Evaluation}

\textbf{Research Questions.}
We organize the experimental evaluation around the following five research questions:

\begin{itemize}
    \item \textbf{RQ1 -- Overall Effectiveness:}
    How effective is GraphCompliance across regulatory settings and encoder backbones?

    \item \textbf{RQ2 -- Accuracy Bottleneck Analysis:}
    Are graph construction and candidate CU retrieval major accuracy bottlenecks in GraphCompliance?

    \item \textbf{RQ3 -- Graph Alignment Effectiveness:}
    How much does Policy--Context Graph alignment contribute to prediction performance?

    \item \textbf{RQ4 -- Handling Regulatory Heterogeneity:}
    Can GraphCompliance accommodate differences in terminology and supervision across regulatory settings?

    \item \textbf{RQ5 -- False-Negative Recovery and Evidence Alignment:}
    What types of positive cases does GraphCompliance recover from text-based representations, and what policy--context alignment patterns are associated with these recoveries?
\end{itemize}

\textbf{Datasets.}
We evaluate GraphCompliance on three public benchmarks---EU2UK, GDPR-13, and CONTRACT---which capture different forms of regulation--context relations. Although all three are formulated as binary prediction tasks in our experiments, their positive labels represent different relations and should not be interpreted as an identical notion of legal compliance.

EU2UK is a regulatory document-to-document retrieval benchmark for identifying UK legislation that transposes a given EU directive~\cite{chalkidis2021doc2doc}. An EU directive serves as the query, while relevant UK documents are defined by official transposition relations. The original benchmark contains a retrieval pool of 52,515 UK legislative documents and provides 1,400 EU directives for training, together with separate development and test queries. A characteristic difficulty of the benchmark is that semantically similar legislation may nevertheless have different relevance labels, for example when documents concern related or amending legislation without constituting the target transposition relation.
We cast EU2UK as binary classification over EU directive--UK document pairs. In this formulation, the EU directive is converted into the Context Graph, and the UK document is converted into the Policy Graph from which Compliance Units are retrieved. Positive pairs are sampled uniformly at random from the annotated transposition relations, while negative pairs are UK documents without an annotated relation to the directive, selected by retrieval so that they remain topically similar to the query. Negatives are constructed at a positive-to-negative ratio of 1:4, and splits are made at the directive level so that pairs from the same directive do not appear in more than one split. Split statistics are given in Table~\ref{tab:data-splits}.

GDPR-13 is derived from the privacy-policy corpus introduced by Liu et al.~\cite{liu2021notified} for identifying disclosure information required under GDPR Article~13. The original annotation scheme distinguishes ten Article~13 disclosure categories from an \textit{Other} category. In our binary formulation, statements belonging to one of the ten disclosure categories are treated as positive and \textit{Other} statements as negative. Unlike the other two benchmarks, GDPR-13 contains a substantial number of fragments, imperatives, and syntactically incomplete statements that are unfavorable to direct entity--relation extraction. Applying the same extraction procedure without accounting for this input form can substantially reduce Context Graph quality and confound graph-alignment performance with upstream extraction failure. We therefore apply a GPT-4.1-based flattening step only to GDPR-13, rewriting each input into a declarative form intended to preserve its semantic content before Context Graph construction. EU2UK and CONTRACT are processed without this step. To keep the comparison controlled, all reproduced baselines on GDPR-13, including the LLM-based baselines, receive the same flattened inputs.

CONTRACT follows the contractual norm-conflict benchmark lineage introduced by Aires et al.~\cite{aires2018offsets,aires2019contractual}. The task concerns semantic conflicts between normative contractual statements, including cases constructed by modifying normative clauses so that they conflict with the surrounding contractual context. Our local binary formulation follows this lineage and predicts whether the target norm forms the designated conflict relation with its contractual context. Consequently, the positive relation in CONTRACT differs from both legislative correspondence in EU2UK and disclosure identification in GDPR-13. Splits are made at the contract level (Table~\ref{tab:data-splits}).

\textbf{Training.}
The shared applicability encoder is trained on the 1,400 EU2UK training directives. In this source-training stage, the shared encoder and the EU2UK Positive Supervision Filter are optimized jointly using document-level correspondence supervision. The resulting encoder is then frozen. For GDPR-13 and CONTRACT, only the regulation-specific Positive Supervision Filter is adapted, using labeled training examples from each target setting. This design separates reusable policy--context applicability representation from the positive criterion associated with each benchmark. We provide the pretrained shared-encoder checkpoint so that target-specific adaptation can be performed without repeating the source-training stage.

\textbf{Baselines.}
Table~\ref{tab:I} compares GraphCompliance with prior approaches to privacy-policy analysis, norm-conflict detection, and textual entailment. Aires et al.~\cite{aires2018offsets,aires2019contractual} model contractual norm conflicts through semantic representations and vector-space relations, while NeuralConflict~\cite{huang2024neuralconflict} jointly learns conflict identification and semantic relatedness in normative text. Liu et al.~\cite{liu2021notified} classify privacy-policy statements according to GDPR Article~13 disclosure categories. Wehnert et al.~\cite{wehnert2022bert} study legal textual entailment using BERT-based representations, and Sun et al. (2017)~\cite{sun2017inner} use a bidirectional LSTM with inner attention for textual entailment. KCI-TEN~\cite{chen2019knowledge} incorporates structured knowledge into textual-entailment modeling. Sun et al. (2025)~\cite{sun2025compliance} combine regulatory knowledge graphs, event-oriented representations, retrieval, and an LLM-based decision component for compliance checking. They also evaluate EU2UK, GDPR-13, and CONTRACT as binary prediction tasks using MCC, and we follow this evaluation setting for all three benchmarks.

Because the amount of input text available to a model directly affects its performance, we set input conditions so that no baseline receives less information than GraphCompliance. Prior methods are reproduced under our protocol where possible, using the same top-10 candidates with each context--candidate pair truncated to 512 tokens; otherwise, values reported in~\cite{sun2025compliance} are used for EU2UK and CONTRACT and are marked in Table~\ref{tab:I}. Reported values are not used for GDPR-13, where flattened inputs differ from the original setting.

Additionally, to distinguish the contribution of retrieval itself from that of Policy--Context Graph alignment, we include several LLM-based RAG systems as baselines. Flat RAG uses the same BGE-M3 candidate preselection and BGE-reranker-v2-m3 reranking stack as GraphCompliance, followed by either Qwen3-30B~\cite{yang2025qwen3} or GPT-5~\cite{openai2025gpt5systemcard} over the retrieved policy evidence. We also evaluate GraphRAG~\cite{edge2024graphrag} in global mode as a graph-based retrieval baseline. For these LLM-based baselines, the maximum input length is set to cover the longest input in each dataset, so they operate on the complete input without truncation. Their input budget is therefore at least as large as that of GraphCompliance. These comparisons provide retrieval-oriented reference points without the explicit Policy--Context Graph alignment used by GraphCompliance. For GDPR-13, LLM-based baselines receive the same flattened statements as GraphCompliance, so that differences in this setting are not attributable to input preprocessing. All LLM baselines are prompted with a one-shot instruction.

\textbf{Implementation Details.}
The full GraphCompliance configuration uses RoBERTa-large~\cite{liu2019roberta} as the shared applicability encoder. Context Graph entity--relation extraction is performed using GPT-4.1, while policy-guided hypernym mapping uses Qwen3-30B. Candidate CUs are first preselected using BGE-M3~\cite{chen2024m3}, which retains at most 30 candidates by dense similarity, and are then reranked using BGE-reranker-v2-m3~\cite{baai_reranker_v2_m3}. The top \(K=10\) candidates are supplied to the Compliance Gate. The default Local Context Graph view uses an \(n=3\) neighborhood.

Reranker and Compliance Gate inputs are limited to 512 tokens in total, including the contextual representation, CU representation, and special tokens. For dense retrieval, long inputs are divided into 256-token chunks; individual chunk embeddings are normalized, averaged, and normalized again. Optimization uses AdamW~\cite{loshchilov2019adamw} with $\beta = (0.9, 0.999)$, $\epsilon = 10^{-8}$, and weight decay $0.01$. The base learning rate is \(10^{-5}\), followed by the selected continuation stage at \(2\times10^{-6}\). The bag batch size is 2, the pair microbatch size is 8, and evaluation processes 32 pairs per batch. Model weights are maintained in FP32 with BF16 autocast. All configuration choices, including K, the neighborhood size n, and decision thresholds, were selected on the development sets. Additional details on decision-threshold calibration and empty-candidate handling are provided in Appendix~\ref{app:training}.

\textbf{Evaluation Metric.}
Because the benchmark distributions are imbalanced, particularly in CONTRACT, we use the Matthews correlation coefficient (MCC) as the primary evaluation metric~\cite{chicco2020mcc}:

\begin{equation}
\label{eq:mcc}
\mathrm{MCC} =
\frac{TP\cdot TN-FP\cdot FN}
{\sqrt{(TP+FP)(TP+FN)(TN+FP)(TN+FN)}}.
\end{equation}

Unless otherwise stated, reported GraphCompliance MCC values are averaged over ten runs with different random seeds. We report MCC separately for each regulatory setting. Cross-setting means are used only as descriptive summaries of configuration-level behavior and are not interpreted as a single aggregate measure over identical label semantics.

\subsection{Overall Effectiveness (RQ1)}
\label{sec:rq1}

RQ1 asks: \textit{How effective is GraphCompliance across regulatory settings and encoder backbones?}

GraphCompliance is designed to provide a common policy--context alignment mechanism across regulatory tasks with different document structures and supervision criteria. We therefore first evaluate the same GraphCompliance configuration on EU2UK, GDPR-13, and CONTRACT. Performance is compared with the literature-based methods and LLM-based RAG baselines described above. This experiment examines whether the proposed formulation remains effective across legislative correspondence, privacy-disclosure identification, and contractual norm-conflict detection, despite the different positive relations defined by the three benchmarks.


\begin{table}[t]
\centering
\caption{MCC comparison on EU2UK, GDPR-13, and CONTRACT. $^{\ast}$ denotes results reported in [4], $^{\dagger}$ denotes results reproduced in this study, and “—” indicates that no result is included in the corresponding comparison setting.}
\label{tab:I}
\begingroup
\small
\setlength{\tabcolsep}{0.5pt}
\renewcommand{\arraystretch}{1.15}
\begin{tabularx}{\linewidth}{@{}>{\raggedright\arraybackslash}XDDD@{}}
\toprule
\shortstack[l]{Method} & \gcdatasethead{EU2UK} & \gcdatasethead{GDPR-13} & \gcdatasethead{CONTRACT} \\
\midrule
Aires et al.~\cite{aires2018offsets}$^{\ast}$ & --- & --- & 0.648 \\
NeuralConflict \cite{huang2024neuralconflict}$^{\ast}$ & --- & --- & 0.661 \\
Liu et al.~\cite{liu2021notified}$^{\dagger}$ & 0.711 & 0.588 & 0.658 \\
Wehnert et al.~\cite{wehnert2022bert}$^{\dagger}$ & 0.705 & 0.611 & 0.686 \\
Sun et al.~(2017) \cite{sun2017inner}$^{\dagger}$ & 0.729 & 0.638 & 0.715 \\
KCI-TEN \cite{chen2019knowledge}$^{\ast}$ & 0.730 & --- & 0.713 \\
Sun et al.~(2025) \cite{sun2025compliance}$^{\ast}$ & 0.778 & --- & 0.730 \\
\midrule
Flat RAG (Qwen3-30B)$^{\dagger}$ & 0.682 & 0.591 & 0.629 \\
Flat RAG (GPT-5)$^{\dagger}$ & 0.744 & 0.612 & 0.698 \\
GraphRAG \cite{edge2024graphrag}$^{\dagger}$ & 0.739 & 0.611 & 0.725 \\
\midrule
GraphCompliance & 0.802{\scriptsize$\pm$0.005} & 0.682{\scriptsize$\pm$0.013} & 0.793{\scriptsize$\pm$0.017} \\
\bottomrule
\end{tabularx}
\endgroup
\end{table}


\begin{table}[t]
\begingroup
\makeatletter
\renewenvironment{figure}[1][]{\def\@captype{figure}}{}
\makeatother
\begin{figure}[t]
\centering
\includegraphics[width=\linewidth]{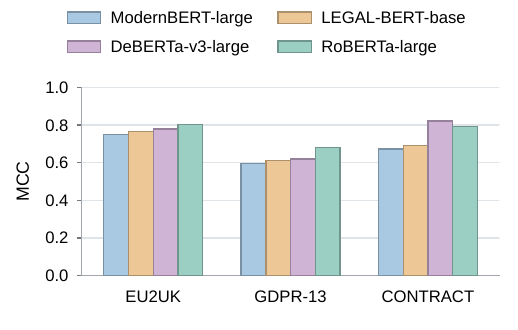}
\caption{MCC by encoder backbone on EU2UK, GDPR-13, and CONTRACT under the same GraphCompliance configuration. The bars reproduce the values in Table~\ref{tab:II} and use a zero-based vertical axis.}
\label{fig:encoder-comparison}
\end{figure}

\endgroup
\par\vspace{12pt}
\begingroup
\renewenvironment{table}[1][]{}{}
\begin{table}[t]
\centering
\caption{Encoder-backbone comparison under the same GraphCompliance Policy Graph, Context Graph, retrieval, and alignment pipeline. Values are MCC; Mean is the arithmetic mean across EU2UK, GDPR-13, and CONTRACT.}
\label{tab:II}
\begingroup
\small
\setlength{\tabcolsep}{0.5pt}
\renewcommand{\arraystretch}{1.15}
\begin{tabularx}{\linewidth}{@{}>{\raggedright\arraybackslash}XDDDD@{}}
\toprule
\shortstack[l]{Encoder} & \gcdatasethead{EU2UK} & \gcdatasethead{GDPR-13} & \gcdatasethead{CONTRACT} & \gcdatasethead{Mean} \\
\midrule
ModernBERT & 0.751 & 0.594 & 0.672 & 0.672 \\
LEGAL-BERT & 0.765 & 0.611 & 0.692 & 0.689 \\
DeBERTa-v3 & 0.779 & 0.620 & 0.821 & 0.740 \\
RoBERTa & 0.802 & 0.682 & 0.793 & 0.759 \\
\bottomrule
\end{tabularx}
\endgroup
\end{table}

\endgroup
\end{table}

\begin{table}[t]
\begingroup
\renewenvironment{table}[1][]{}{}
\begin{table}[t]
\centering
\caption{Cycle-consistency of Policy and Context Graph construction over five text-to-graph-to-text iterations. Semantic columns report reconstruction similarity, while Structural columns report graph-level structural consistency relative to the initial graph.}
\label{tab:IX}
\begingroup
\small
\setlength{\tabcolsep}{2.5pt}
\renewcommand{\arraystretch}{1.15}
\begin{tabularx}{\linewidth}{@{}c*{4}{>{\centering\arraybackslash}X}@{}}
\toprule
Iteration & \multicolumn{2}{c}{Policy Graph} & \multicolumn{2}{c}{Context Graph} \\
\cmidrule(lr){2-3}\cmidrule(lr){4-5}
 & Semantic & Structural & Semantic & Structural \\
\midrule
1 & 0.925 & 1.000 & 0.891 & 0.892 \\
2 & 0.901 & 1.000 & 0.873 & 0.882 \\
3 & 0.896 & 1.000 & 0.868 & 0.871 \\
4 & 0.889 & 1.000 & 0.865 & 0.880 \\
5 & 0.889 & 1.000 & 0.854 & 0.875 \\
\bottomrule
\end{tabularx}
\endgroup
\end{table}

\par\vspace{14pt}
\begin{table}[t]
\centering
\caption{Sensitivity of reconstruction-based graph fidelity to controlled noise injection. The corruption level \(\delta\) increases edge and attribute perturbation, and the reported value is mean semantic similarity between the source text and reconstruction from the corrupted graph.}
\label{tab:X}
\begingroup
\small
\setlength{\tabcolsep}{8pt}
\renewcommand{\arraystretch}{1.15}
\begin{tabular}{@{}cc@{}}
\toprule
Corruption level \(\delta\) & Mean semantic similarity \\
\midrule
0.00 & 0.891 \\
0.03 & 0.865 \\
0.05 & 0.851 \\
0.10 & 0.818 \\
0.20 & 0.764 \\
\bottomrule
\end{tabular}\par
\endgroup
\end{table}

\endgroup
\end{table}

\begin{table}[t]
\centering
\caption{Retrieval-budget sensitivity on EU2UK and CONTRACT. \(K\) is the number of reranked candidate CUs supplied to the Compliance Gate, values are MCC, and ``All'' supplies the complete candidate set.}
\label{tab:III}
\begingroup
\small
\setlength{\tabcolsep}{0.5pt}
\renewcommand{\arraystretch}{1.15}
\begin{tabular}{@{}>{\centering\arraybackslash}p{18pt}DD@{}}
\toprule
\(K\) & \gcdatasethead{EU2UK} & \gcdatasethead{CONTRACT} \\
\midrule
1 & 0.713 & 0.743 \\
3 & 0.744 & 0.793 \\
5 & 0.787 & 0.793 \\
10 & 0.802 & 0.793 \\
20 & 0.802 & 0.723 \\
All & 0.798 & 0.723 \\
\bottomrule
\end{tabular}\par
\endgroup
\end{table}

\begin{table}[t]
\centering
\caption{Weak-target candidate-retrieval coverage on the 234 positive development instances of GDPR-13. ``Positive bags covered'' counts positive instances containing at least one mapped target CU, ``Any-target coverage'' is the corresponding bag-level proportion, and ``Weak unit recall'' is recall over mapped target CUs. The mappings are used as weak retrieval targets rather than independently annotated CU-level gold.}
\label{tab:IV}
\begingroup
\small
\setlength{\tabcolsep}{5pt}
\renewcommand{\arraystretch}{1.15}
\begin{tabular}{@{}>{\centering\arraybackslash}p{18pt}ccc@{}}
\toprule
\(K\) & \shortstack{Positive bags\\covered} & \shortstack{Any-target\\coverage} & \shortstack{Weak unit\\recall} \\
\midrule
1 & 92 / 234 & 0.3932 & 0.2110 \\
3 & 164 / 234 & 0.7009 & 0.4358 \\
5 & 194 / 234 & 0.8291 & 0.6009 \\
10 & 232 / 234 & 0.9915 & 0.9060 \\
20 & 234 / 234 & 1.0000 & 1.0000 \\
All & 234 / 234 & 1.0000 & 1.0000 \\
\bottomrule
\end{tabular}\par
\endgroup
\end{table}

\begin{table}[t]
\begingroup
\makeatletter
\renewenvironment{figure}[1][]{\def\@captype{figure}}{}
\makeatother
\begin{figure}[t]
\centering
\includegraphics[width=\linewidth]{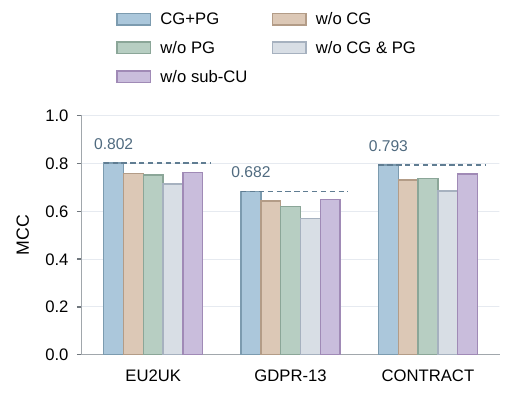}
\caption{MCC under the full CG+PG representation and ablations on EU2UK,
GDPR-13, and CONTRACT. ``w/o CG'' replaces the Context Graph with raw
context, ``w/o PG'' replaces the Policy Graph with raw policy text,
``w/o CG \& PG'' uses raw text for both, and ``w/o sub-CU'' disables
structural CU expansion while retaining both graphs; dashed lines mark
each dataset's full CG+PG score.}
\label{fig:graph-ablation}
\end{figure}

\endgroup
\end{table}

\begin{table}[t]
\centering
\caption{EU2UK diagnostic stratified by original Context Graph size. \(G_C\) and \(T_C\) denote MCC using the Context Graph and raw context, respectively, with the Policy Graph retained in both conditions; \(G_C-T_C\) is the within-stratum difference.}
\label{tab:VI}
\begingroup
\small
\setlength{\tabcolsep}{8pt}
\renewcommand{\arraystretch}{1.15}
\begin{tabular}{@{}cccc@{}}
\toprule
\# CG nodes & \(G_C\) & \(T_C\) & \(G_C-T_C\) \\
\midrule
\(\leq30\) & 0.903 & 0.881 & +0.022 \\
\(>30\) & 0.719 & 0.645 & +0.074 \\
\bottomrule
\end{tabular}\par
\endgroup
\end{table}

\begin{table}[t]
\centering
\caption{Policy-guided hypernym ablation on EU2UK and GDPR-13. Values are MCC for no hypernym mapping, hypernym mapping without policy grounding, and policy-guided hypernym mapping; CONTRACT is omitted because its policy representation does not provide a comparable premise set for this analysis.}
\label{tab:VII}
\begingroup
\small
\setlength{\tabcolsep}{0.5pt}
\renewcommand{\arraystretch}{1.15}
\begin{tabularx}{\linewidth}{@{}>{\raggedright\arraybackslash}XDD@{}}
\toprule
\shortstack[l]{Setting} & \gcdatasethead{EU2UK} & \gcdatasethead{GDPR-13} \\
\midrule
No hypernym & 0.782 & 0.655 \\
Hypernym without policy grounding & 0.780 & 0.662 \\
Policy-guided hypernym & 0.802 & 0.682 \\
\bottomrule
\end{tabularx}
\endgroup
\end{table}

\begin{table}[t]
\centering
\caption{Comparison of representation sharing and regulation-specific adaptation. Values are MCC for a shared encoder with one common head, three independent encoders, and the shared encoder with regulation-specific Positive Supervision Filters; the shared encoder is pretrained on EU2UK before target-specific adaptation.}
\label{tab:VIII}
\begingroup
\small
\setlength{\tabcolsep}{0.5pt}
\renewcommand{\arraystretch}{1.15}
\begin{tabularx}{\linewidth}{@{}>{\raggedright\arraybackslash}XDDD@{}}
\toprule
\shortstack[l]{Configuration} & \gcdatasethead{EU2UK} & \gcdatasethead{GDPR-13} & \gcdatasethead{CONTRACT} \\
\midrule
Shared encoder + one head & 0.785 & 0.582 & 0.776 \\
Three independent encoders & 0.802 & 0.678 & 0.799 \\
Shared encoder + Positive Supervision Filter & 0.802 & 0.682 & 0.793 \\
\bottomrule
\end{tabularx}
\endgroup
\end{table}

With RoBERTa-large as the shared applicability encoder, GraphCompliance achieves MCCs of 0.802, 0.682, and 0.793 on EU2UK, GDPR-13, and CONTRACT, respectively. The highest MCCs among the prior methods in Table~\ref{tab:I} are 0.778, 0.638, and 0.730, corresponding to differences of 0.024, 0.044, and 0.063. GraphCompliance is also numerically higher than the additional Flat RAG and GraphRAG baselines in all three settings. The results show that a single GraphCompliance formulation can be applied across the three benchmark relations while maintaining competitive predictive performance.

We next examine whether the observed performance depends on the choice of shared applicability encoder. RoBERTa-large, DeBERTa-v3-large~\cite{he2021debertav3}, ModernBERT-large~\cite{warner2025modernbert}, and LEGAL-BERT-base~\cite{chalkidis2020legalbert} are evaluated while retaining the same Policy Graph, Context Graph, candidate-retrieval pipeline, and downstream alignment procedure. Only the encoder backbone is changed. This experiment therefore tests whether the proposed alignment interface can be used with different pretrained encoder families rather than requiring an architecture selected specifically for each regulatory setting (Table~\ref{tab:II} and Figure~\ref{fig:encoder-comparison}).

RoBERTa-large achieves MCCs of 0.802, 0.682, and 0.793 on EU2UK, GDPR-13, and CONTRACT, respectively, yielding the highest cross-setting mean MCC of 0.759 among the evaluated backbones. DeBERTa-v3-large achieves 0.779 on EU2UK and the highest CONTRACT score of 0.821, while the remaining encoder families also operate within the same GraphCompliance formulation. No single backbone is best on every benchmark. These results indicate that the alignment interface is not specific to a single encoder architecture. Throughout the remaining experiments, we report RoBERTa-large as the main configuration so that a single encoder setting is used consistently across the three benchmarks.

\subsection{Accuracy Bottleneck Analysis (RQ2)}
\label{sec:rq2}

RQ2 asks: \textit{Are graph construction and candidate CU retrieval major accuracy bottlenecks in GraphCompliance?}

\textbf{Graph Fidelity.} GraphCompliance does not feed source text directly into the prediction model; instead, the Policy Graph and Context Graph serve as intermediate representations for subsequent policy--context alignment. Information lost or distorted during graph construction could therefore limit downstream performance independently of the Compliance Gate. To assess this possibility, we propose two complementary proxy evaluations of graph fidelity: the Cycle-Consistency Test and the Noise-Injection Test. The Cycle-Consistency Test begins with a single reconstruction \(T_0\rightarrow G_0\rightarrow T_1\) and repeatedly applies \(T_k\rightarrow G_k\rightarrow T_{k+1}\), measuring whether the information represented in the graph remains stable through successive text-to-graph and graph-to-text transformations. We evaluate this stability in terms of semantic consistency between the source and reconstructed text and structural consistency between graph-level statistics of the initial and reconstructed graphs. Cycle consistency alone, however, may remain high even when the same omission or distortion is reproduced across iterations. We therefore complement it with the Noise-Injection Test, which deliberately perturbs the graph through edge deletion, spurious-edge addition, and CU-attribute corruption and measures the resulting change in reconstruction similarity. The two tests address complementary properties: cycle consistency evaluates whether the constructed representation preserves source information under repeated reconstruction, whereas noise injection verifies that the fidelity measure is sufficiently sensitive to actual degradation of graph structure and attributes.

Both graph representations remain stable under repeated reconstruction. For the Policy Graph, semantic consistency decreases from 0.925 after the first reconstruction to 0.889 after the fifth, while structural consistency remains 1.000 throughout. The Context Graph shows a gradual decrease in semantic consistency from 0.891 to 0.854, while structural consistency remains between 0.892 and 0.871 (Table~\ref{tab:IX}). In the Noise-Injection Test, semantic similarity decreases monotonically from 0.891 at \(\delta=0.00\) to 0.764 at \(\delta=0.20\), indicating that the fidelity measure responds systematically as structural and attribute corruption increases (Table~\ref{tab:X}). Taken together, the high reconstruction consistency and the monotonic response to controlled corruption indicate that the Policy Graph and Context Graph preserve the principal semantic and structural information of their source inputs under the evaluated procedure. These proxy evaluations therefore provide evidence of representation stability, although they do not rule out systematic graph-construction errors or their contribution to individual downstream prediction failures.



\textbf{Candidate Retrieval Coverage.} Even when the source information is faithfully represented in the graphs, the Compliance Gate can evaluate only the CUs exposed by the retrieval stage. An overly narrow candidate set may omit relevant regulatory requirements, whereas an overly broad set may introduce low-relevance CUs and make downstream discrimination more difficult. We therefore vary the retrieval budget over \(K\in\{1,3,5,10,20,\mathrm{All}\}\) on EU2UK and CONTRACT and measure the resulting end-to-end MCC. This experiment evaluates whether prediction remains sensitive to the breadth of the retrieved candidate set and whether the default retrieval depth is sufficient to expose useful policy evidence without unnecessarily increasing the downstream comparison space.

On EU2UK, MCC increases from 0.713 at \(K=1\) to 0.802 at \(K=10\), after which it remains nearly saturated at 0.802 for \(K=20\) and 0.798 when all CUs are supplied. CONTRACT increases from 0.743 at \(K=1\) to 0.793 at \(K=3\), maintains the same performance through \(K=10\), and decreases to 0.723 at \(K=20\) and \(\mathrm{All}\). Increasing the candidate pool therefore improves performance when the initial retrieval budget is restrictive, but exposing the gate to additional low-relevance CUs does not continue to improve prediction. The retrieval budget \(K = 10\) was selected on the development set; Table~\ref{tab:III} reports the corresponding test-set sensitivity.

End-to-end saturation with respect to \(K\) does not directly establish that CUs associated with positive instances are actually present among the retrieved candidates. We therefore perform an additional coverage analysis on GDPR-13, whose Article 13 disclosure categories allow weak CU-level target mappings to be constructed. For each positive instance, any-target coverage measures whether at least one mapped target CU is included in the retrieved candidate set, while weak-unit recall measures the fraction of mapped target CUs that are retrieved. These mappings are used only as weak retrieval targets for coverage analysis and are not treated as independently annotated CU-level legal gold.

At \(K=10\), 232 of 234 development instances contain at least one mapped target CU, corresponding to an any-target coverage of 0.991, while weak-unit recall reaches 0.906. Among the 44 false negatives on these instances, 43 occur in instances for which a mapped target CU is already present among the retrieved candidates (Table~\ref{tab:IV}). This indicates that most of the observed false negatives are not explained by failure to expose any mapped relevant CU to the Compliance Gate. Together with the retrieval-budget experiment, the coverage analysis shows that the default \(K=10\) configuration provides high candidate availability on GDPR-13. Together with the graph-fidelity results, these findings motivate prioritizing evidence discrimination within the Compliance Gate for further performance improvement under the evaluated configuration.

\subsection{Graph Alignment Effectiveness (RQ3)}
\label{sec:rq3}

RQ3 asks: \textit{How much does Policy--Context Graph alignment contribute to prediction performance?}

The central design of GraphCompliance is to represent regulatory requirements as a Policy Graph and the assessed situation as a Context Graph, and then align the two representations in the Compliance Gate. To examine whether these structured representations contribute to prediction performance, we conduct a controlled representation ablation in which one or both representations are replaced with their raw-text counterparts. Raw context + PG retains the Policy Graph while replacing the Context Graph with the source context, CG + raw policy retains the Context Graph while replacing the Policy Graph with raw policy text, and raw context + raw policy removes both graph representations. The full CG + PG configuration provides the reference condition. When the Policy Graph is replaced, the raw policy text is divided into fixed-length chunks, which take the place of Compliance Units in the same candidate-retrieval and reranking pipeline. To avoid evaluating a model on an input form it was not trained on, the shared applicability encoder is retrained from the same RoBERTa-large initialization for each configuration using the corresponding input form, followed by the same target-specific Positive Supervision Filter adaptation and development-set threshold calibration. The encoder backbone, training data, optimization settings, and MIL decision procedure are otherwise kept identical across configurations. We additionally evaluate a w/o structural CU expansion condition that retains both graph representations and the same semantic candidate retrieval and reranking, but disables the deterministic Policy Graph lookup used to recover referenced and structurally related governing or subordinate CUs.

The complete CG + PG configuration achieves MCCs of 0.802, 0.682, and 0.793 on EU2UK, GDPR-13, and CONTRACT, respectively, with a mean of 0.759. Replacing the Context Graph with raw context reduces the scores to 0.778, 0.662, and 0.751, yielding a mean of 0.730. Replacing the Policy Graph with raw policy text produces 0.771, 0.640, and 0.756, with a mean of 0.722. When both graph representations are replaced with raw text, performance further decreases to 0.755, 0.611, and 0.726, with a mean of 0.697. Relative to the full configuration, removing the Context Graph decreases mean MCC by 0.029, removing the Policy Graph decreases it by 0.037, and removing both decreases it by approximately 0.062. The decline occurs across all three benchmarks, while the largest overall decrease is observed when neither structured representation is used. The decrease from removing both representations (0.062) is close to the sum of the individual decreases (0.029 and 0.037), indicating that the Policy Graph and Context Graph contribute largely independent information to policy--context alignment rather than serving as interchangeable representations (Figure~\ref{fig:graph-ablation}). Disabling structural CU expansion while retaining both graph representations yields MCCs of 0.761, 0.648, and 0.756 on EU2UK, GDPR-13, and CONTRACT, respectively, with a mean of 0.722, corresponding to a mean decrease of 0.037 from the full configuration. Because the Policy Graph and Context Graph remain available in this condition, the decrease suggests that Policy Graph connectivity contributes beyond the structured CU representation itself by exposing regulatory dependencies that may not be selected through semantic retrieval alone.

We further examine whether the contribution of the Context Graph changes as the extracted contextual representation becomes larger. In Local mode, GraphCompliance selects the neighborhood around actors and events relevant to a candidate CU rather than supplying the complete Context Graph uniformly. As the number of extracted entities and relations increases, the relevant actions, conditions, and participating actors are distributed across a larger contextual representation, potentially increasing the value of this candidate-conditioned selection. We therefore stratify the EU2UK development population according to the number of nodes in the original Context Graph, using groups with at most 30 nodes and more than 30 nodes, and compare the graph-based representation with raw context + PG.

For Context Graphs containing at most 30 nodes, the graph-based configuration achieves an MCC of 0.903, compared with 0.881 for raw context + PG, a difference of 0.022. For graphs containing more than 30 nodes, the corresponding scores are 0.719 and 0.645, increasing the difference to 0.074. Although absolute performance is lower in the larger-graph group for both representations, the relative benefit of the Context Graph becomes more pronounced as the extracted contextual representation grows larger. This result is consistent with the role of Local mode in selecting the actors, actions, conditions, and surrounding evidence most relevant to the candidate regulatory unit rather than requiring the encoder to process all contextual information uniformly (Table~\ref{tab:VI}).

Together, the representation ablation, structural-expansion ablation, and size-stratified analysis show that both the Policy Graph and Context Graph contribute to prediction performance. Their contributions are largely independent, the Policy Graph benefit depends substantially on recovering structurally related provisions, and the relative benefit of the Context Graph is greater when the extracted contextual representation contains a larger number of entities and relations.

\subsection{Handling Regulatory Heterogeneity (RQ4)}
\label{sec:rq4}

RQ4 asks: \textit{Can GraphCompliance accommodate differences in terminology and positive supervision across regulations?}

Regulatory text and real-world context can express the same underlying concept using substantially different terminology. GraphCompliance addresses this mismatch through policy-guided hypernym mapping, which associates Context Graph entities with higher-level concepts grounded in the target regulation. To distinguish the effect of policy grounding from that of semantic abstraction alone, we compare three conditions: No hypernym, Hypernym without policy grounding, and Policy-guided hypernym. This analysis is conducted on EU2UK and GDPR-13, whose policy representations provide premise sets suitable for policy-guided hypernym mapping.

The policy-guided configuration achieves MCCs of 0.802 on EU2UK and 0.682 on GDPR-13 (Table~\ref{tab:VII}). Removing hypernyms decreases the scores to 0.782 and 0.655, while using hypernyms without policy grounding yields 0.780 and 0.662. Policy-guided mapping improves over no hypernyms by 0.020 on EU2UK and 0.027 on GDPR-13, and over ungrounded hypernyms by 0.022 and 0.020, respectively. Ungrounded hypernyms are not consistently beneficial: they slightly reduce performance on EU2UK and provide only a limited improvement on GDPR-13. These results support the importance of grounding higher-level concepts in the terminology and definitions of the target regulation rather than introducing semantic abstractions independently of the policy.

Regulatory heterogeneity also arises in the positive boundary of a policy--context relation. As illustrated by the motivating example in Fig.~\ref{fig:failure-modes}, similar actions may lead to different assessments when the governing scope, actor roles, definitions, conditions, or exceptions differ across regulations. The specificity required for positive evidence can therefore vary even when the surface semantics of the policy--context relation are similar. GraphCompliance addresses this variation by separating a shared applicability encoder from a regulation-specific Positive Supervision Filter. The shared encoder learns a reusable representation of policy--context applicability, whereas the filter adapts the resulting representation to the positive criterion of the target regulation. We evaluate this design against two alternatives under the same MIL setting. Independent three encoders trains a separate encoder and regulation-specific classification head for each regulation using the same MIL objective and aggregation, whereas Shared Encoder + One Head shares both the applicability 
representation and the positive decision layer across regulations. These comparisons separate the effect of regulation-specific adaptation from the choice of MIL aggregation, which is held fixed across the three configurations.

GraphCompliance achieves MCCs of 0.802, 0.682, and 0.793 on EU2UK, GDPR-13, and CONTRACT, respectively, with a mean of 0.759. Independent encoders trained under the same MIL setting yield 0.802, 0.678, and 0.799, with a mean of 0.760. On EU2UK, the two configurations coincide by construction, because the shared encoder is pretrained on EU2UK and paired with the EU2UK-specific filter. On the two target settings not used during shared-encoder pretraining, the shared configuration is 0.004 higher on GDPR-13 and 0.006 lower on CONTRACT than the independently fine-tuned encoders. Sharing both the encoder and a single common decision head results in lower MCCs of 0.785, 0.582, and 0.776, respectively, with a mean of 0.714 and the largest decrease on GDPR-13 (Table~\ref{tab:VIII}).

These results indicate that the shared encoder retains a reusable representation of policy--context applicability when combined with regulation-specific adaptation. Because GDPR-13 and CONTRACT are not used during shared-encoder pretraining, the comparable observed performance suggests that the learned applicability representation remains useful beyond the regulation on which it was trained. Because the independent alternative uses the same MIL setting and differs mainly in fine-tuning a separate encoder for each regulation, this comparison indicates that encoder sharing incurs little cost on the target regulations. In contrast, sharing a single common decision head substantially reduces performance, particularly on GDPR-13, supporting the need for regulation-specific positive supervision.

\subsection{False-Negative Recovery and Evidence Alignment (RQ5)}
\label{sec:rq5}

RQ5 asks: \textit{What types of positive cases does GraphCompliance recover from text-based representations, and what policy--context alignment patterns are associated with these recoveries?}

Aggregate performance and representation ablations establish whether graph-based representations improve prediction, but they do not reveal which types of evidence become easier to recognize through policy--context alignment. We therefore conduct a qualitative, sample-level analysis of recovered instances using saved development-set predictions from a single run of the final GraphCompliance configuration and the TextOnly and NoCG controls, for which sample-level predictions can be linked to the corresponding model inputs. A recovery is defined, separately for each control, as a positive instance that the control predicts as negative but GraphCompliance correctly identifies. TextOnly uses raw context and raw policy text, whereas NoCG retains the Policy Graph while replacing the Context Graph with raw context. Because the analysis is performed on a single run and examines recovered instances only, it characterizes the evidence patterns associated with recovery rather than estimating net performance gains; positives identified by a control but missed by GraphCompliance are not analyzed here, and the aggregate effect of each representation is given by the ablation in RQ3.

\textbf{Scope of the analyzed recoveries.} Relative to TextOnly, GraphCompliance recovers 9 positive instances on EU2UK and 30 on GDPR-13. Relative to NoCG, it recovers 24 on EU2UK and 37 on GDPR-13. These instances form the population examined below.

\textbf{Purpose and temporal qualification.} The clearest concentration appears in GDPR-13. Of the 30 instances recovered relative to TextOnly, 16 belong to Data Processing Purposes and 7 to Data Retention Period. Together they account for 23 of the 30 recoveries (76.7\%), although these two categories represent only 103 of the 234 positive instances (44.0\%) in the analyzed population. The same over-representation appears relative to NoCG, where purpose- and retention-related instances account for 24 of 37 recoveries (64.9\%). In these categories, positive evidence depends on relating an action to its purpose, condition, or temporal qualification rather than on an isolated lexical cue. In one retention-period example, the context states that a retention or documentation period is ``2 to 10 years.'' The Context Graph explicitly connects the retention concept with its temporal value, making the relation directly comparable with the corresponding retention-period requirement. In another example, no fixed duration is stated; instead, the data are deleted when they are ``no longer necessary'' to retain. The Context Graph represents this statement as a relation between the deletion action and its retention condition. Although the surface expressions differ, both are organized into the same functional relation between a data action and its temporal or conditional qualification.

\textbf{Document relationships.} In EU2UK, recovered instances include cases in which the relevant evidence lies in relations between legislative documents rather than in shared terminology. In one example, explicit amendment relations between an EU directive and earlier directives organize a long legislative context around the document relationships relevant to the corresponding UK provision.

Across these patterns, the recurring advantage is not the preservation of regulatory terminology but the organization of contextual evidence into relations---purpose, condition, temporal qualification, document relationship---that can be aligned with structured policy evidence. The analysis therefore complements the aggregate ablation in RQ3 by indicating where graph alignment operates at the sample level: recovery is most frequent when positive evidence is distributed across such relations rather than expressed through a single lexical cue.
\section{Limitations}

GraphCompliance depends on the quality of Policy Graph and Context Graph construction. Errors in extracted entities, relations, conditions, or regulatory dependencies can remove or distort evidence before policy--context alignment. This limitation is more pronounced for fragmentary or syntactically irregular inputs. The reconstruction-based analyses in RQ2 provide evidence of representation stability, but do not replace evaluation against independently annotated reference graphs.


GraphCompliance also requires labeled supervision when adapting to a new regulation. Although the shared applicability encoder can be reused, a regulation-specific Positive Supervision Filter must still be fitted using labeled examples from the target regulation; the present experiments use the labeled training splits of GDPR-13 and CONTRACT. Moreover, the present evaluation covers three regulatory settings. Broader experiments are therefore needed to assess how well the shared applicability representation transfers to regulations with substantially different structures, terminology, and jurisdictions.
\section{Conclusion}

This paper presented GraphCompliance, a structure-aware framework for regulatory compliance-related prediction based on alignment between Policy and Context Graphs. Regulatory requirements are represented as structured Compliance Units within a Policy Graph that preserves their normative fields and structural dependencies, while contextual information is represented as entity--relation structures in a Context Graph. Policy-guided hypernym mapping connects contextual entities to higher-level concepts grounded in the target regulation, and the Compliance Gate aligns the resulting policy and contextual evidence through a shared applicability encoder and a regulation-specific Positive Supervision Filter.

GraphCompliance achieves MCCs of 0.802, 0.682, and 0.793 on EU2UK, GDPR-13, and CONTRACT, respectively, and is numerically higher than the compared methods in all three regulatory settings. Representation ablations show that replacing either the Policy Graph or the Context Graph with raw text reduces performance, with the largest decrease occurring when both structured representations are removed. The policy-guided hypernym analysis further shows higher performance than both the no-hypernym and ungrounded-hypernym conditions on EU2UK and GDPR-13. Together, these results indicate that structured representations of both regulatory requirements and contextual evidence contribute to policy--context alignment.

The representation-sharing experiments provide an additional result. A shared applicability encoder pretrained on EU2UK and combined with regulation-specific Positive Supervision Filters performs comparably to independently fine-tuned encoders under the same MIL setting, differing by at most 0.006 MCC on the target settings. At the same time, sharing both the encoder and a single common decision filter reduces performance. These observations suggest that the shared encoder captures a reusable representation of policy--context applicability that remains useful in regulatory settings not observed during its pretraining, while the positive decision boundary still benefits from adaptation to the target regulation.

Overall, the results support separating reusable policy--context applicability representation from regulation-specific decision adaptation. The Policy Graph preserves regulatory structure, the Context Graph organizes distributed contextual evidence, and their alignment provides a common interface through which these two forms of information can be compared. Future work will examine graph construction for more irregular inputs, evaluate graph fidelity against independently annotated reference graphs, reduce reliance on external preprocessing, and study the transfer of the shared applicability representation across additional regulations, jurisdictions, and deployment settings.

\section*{Acknowledgment}
Generative AI tools (ChatGPT and Codex) were used to assist with code generation and code review, as well as initial drafting and translation of text in Sections II (Related Work), III (Methodology), and IV (Experimental Evaluation). All AI-assisted text and all AI-generated code used in this study were reviewed and validated by the authors.
\gcappendixlayouttrue
\setcounter{topnumber}{2}
\setcounter{totalnumber}{2}
\renewcommand{\topfraction}{0.9}
\renewcommand{\textfraction}{0.08}
\makeatletter
\setlength{\@fpsep}{18pt plus 2pt minus 2pt}
\setlength{\@dblfpsep}{18pt plus 2pt minus 2pt}
\makeatother
\appendices
\appendices

\section{Data, Training, and Reproducibility}
\label{app:training}

\subsection{Dataset Splits}
\label{app:splits}

Table~\ref{tab:data-splits} summarizes the binary formulations used in
our experiments. EU2UK pairs are constructed from the 1,400 training
directives and the separate development and test directives of the
original benchmark at a positive-to-negative ratio of 1:4, with splits
made at the directive level. GDPR-13 statements are drawn from the
corpus of Liu et al.~\cite{liu2021notified}. CONTRACT targets are split
at the contract level, so that no contract contributes targets to more
than one split. Positives are rare in all CONTRACT splits, and the
development split contains only 10 positive targets. Development splits
are used for configuration selection and decision-threshold
calibration, and all main results are reported on the test splits.

\subsection{Training and Inference Configuration}

The final training procedure separates reusable policy--context representation learning from regulation-specific positive decision adaptation. The shared applicability encoder is initialized from RoBERTa-large and trained on the 1,400 EU2UK training directives. During this source-training stage, the shared encoder and the EU2UK Positive Supervision Filter are optimized jointly using document-level EU--UK correspondence supervision. Because this supervision is available at the document-pair level rather than for individual Compliance Units, retrieved CU--Context Graph pairs are treated as a multiple-instance bag instead of assigning every CU in a positive document an independent positive label.

After source training, the shared applicability encoder is frozen. For GDPR-13 and CONTRACT, only the regulation-specific Positive Supervision Filter is optimized, using labeled training examples from the corresponding regulation. The resulting shared-encoder checkpoint is released with the implementation so that regulation-specific adaptation can be performed without repeating the source-training stage.

Context Graph entity--relation extraction is performed using GPT-4.1. For GDPR-13, GPT-4.1 is additionally used for the flattening step described above. Policy-guided hypernym mapping is performed using Qwen3-30B.

\begin{table}[!t]
\centering
\caption{Dataset statistics after binary formulation. Splits are made at
the directive level for EU2UK and at the contract level for CONTRACT,
whose train, development, and test splits contain 6, 2, and 4
contracts, respectively.}
\label{tab:data-splits}
\small
\setlength{\tabcolsep}{4pt}
\begin{tabular*}{\linewidth}{@{\extracolsep{\fill}}llrrrr@{}}
\hline
Dataset & Split & Total & Pos. & Neg. & Pos.\,\% \\
\hline
\multirow{3}{*}{EU2UK}
& Train & 12,550 & 2,510 & 10,040 & 20.00 \\
& Dev   &  1,125 &   225 &    900 & 20.00 \\
& Test  &  1,120 &   224 &    896 & 20.00 \\
\hline
\multirow{3}{*}{GDPR-13}
& Train &  9,774 & 1,429 &  8,345 & 14.62 \\
& Dev   &  2,100 &   234 &  1,866 & 11.14 \\
& Test  &  2,139 &   318 &  1,821 & 14.87 \\
\hline
\multirow{3}{*}{CONTRACT}
& Train &  1,589 &    49 &  1,540 &  3.08 \\
& Dev   &    427 &    10 &    417 &  2.34 \\
& Test  &    984 &    35 &    949 &  3.56 \\
\hline
\end{tabular*}
\end{table}

Candidate CU selection uses a two-stage retrieval pipeline. BGE-M3 first preselects at most 30 CUs by dense similarity, after which BGE-reranker-v2-m3 reranks the candidates. The default configuration retains the top \(K=10\) CUs without an additional score cutoff. The default Context Graph Local mode uses an \(n=3\) neighborhood.

Reranker and Compliance Gate inputs are limited to 512 tokens in total, including the contextual representation, CU representation, and special tokens. For dense retrieval, long inputs are divided into normalized 256-token chunks. The chunk embeddings are averaged and normalized again to form the final dense representation.

Optimization uses AdamW with
\(\beta=(0.9,0.999)\),
\(\epsilon=10^{-8}\),
and weight decay \(0.01\).
The base learning rate is \(10^{-5}\), followed by the selected continuation stage at \(2\times10^{-6}\).
The bag batch size is 2, the pair microbatch size is 8, and evaluation processes 32 pairs per batch. Model weights are maintained in FP32 with BF16 autocast.

For each regulation, the decision threshold is calibrated on development-set bag logits. We exhaustively evaluate the distinct development scores together with the boundary corresponding to all-negative predictions and select the threshold that maximizes development MCC. Ties are resolved in favor of the larger threshold. The selected threshold is then fixed before final evaluation.

Samples without a trainable candidate CU are omitted from gradient updates during training. At evaluation time, these samples remain in the evaluation population and are assigned a fixed bag logit of \(-10^{9}\), after which the same regulation-specific decision threshold is applied.

\begin{table}[t]
\centering
\caption{Inference-time aggregation sensitivity on EU2UK and GDPR-13
using saved development-set logits from a single run of the final
max-MIL model. No retraining or threshold recalibration is performed;
MCC, positive recall, false-negative rate (FNR), and false-positive
rate (FPR) use the original maximum-aggregation threshold.}
\label{tab:aggregation-sensitivity}
\begin{tabular}{llrrrr}
\hline
Setting & Aggregation & MCC & Recall & FNR & FPR \\
\hline
\multirow{3}{*}{EU2UK}
& Maximum & 0.802 & 0.880 & 0.120 & 0.052 \\
& Mean logit & 0.722 & 0.644 & 0.356 & 0.016 \\
& Top-three mean & 0.771 & 0.747 & 0.253 & 0.024 \\
\hline
\multirow{3}{*}{GDPR-13}
& Maximum & 0.683 & 0.812 & 0.188 & 0.057 \\
& Mean logit & 0.561 & 0.581 & 0.419 & 0.042 \\
& Top-three mean & 0.557 & 0.637 & 0.363 & 0.057 \\
\hline
\end{tabular}
\end{table}

\subsection{Sample-level Diagnostics}

Several supplementary analyses require sample-level predictions,
candidate scores, model inputs, and checkpoints to be linked exactly.
For these analyses, we use saved development-set outputs from a single
run of the final GraphCompliance configuration, covering 1,125 EU2UK
pairs and 2,100 GDPR-13 statements. CONTRACT is excluded from these
analyses because its development split contains only 10 positive
targets; it is used only for decision-threshold calibration. These
outputs are used for the aggregation-sensitivity analysis in
Appendix~\ref{app:aggregation}, the candidate-coverage analysis in RQ2,
and the false-negative recovery analysis in RQ5. Because the decision
thresholds are also calibrated on the development sets, these analyses
characterize model behavior rather than estimate test performance, and
their values can differ from the ten-run test averages reported in the
main evaluation.

For RQ5, the stored predictions are joined with their corresponding model inputs using the original evaluation keys, checkpoint, and decision thresholds. No additional training, inference, threshold optimization, score modification, or label modification is performed as part of the recovery analysis.

\section{Aggregation Sensitivity}
\label{app:aggregation}

GraphCompliance uses max-based multiple-instance aggregation over the retrieved CU--context pairs associated with each sample. To examine how the recorded predictions depend on this aggregation choice, we reuse the saved top-10 pair logits from the final max-MIL model and compare three inference-time aggregation rules: maximum, mean logit, and top-three mean logit.

The model is not retrained for this analysis. In addition, the decision threshold originally calibrated for maximum aggregation is retained for all three aggregation rules. The experiment therefore measures inference-time sensitivity of the recorded model outputs rather than comparing independently trained MIL objectives.

Table~\ref{tab:aggregation-sensitivity} reports MCC, positive recall, false-negative rate (FNR), and false-positive rate (FPR). Maximum aggregation yields the highest MCC under the original max-MIL calibration in both regulatory settings. Mean-based aggregation generally reduces the false-positive rate but also lowers positive recall under the unchanged threshold. Because the underlying model and threshold were optimized for maximum aggregation, these results characterize the sensitivity of the existing model rather than establishing a general ordering among alternative MIL training objectives.

\section{Shared-Encoder Resource Efficiency}
\label{app:resource}

We additionally examine the deployment-side resource implications of sharing the applicability encoder. This diagnostic compares one resident RoBERTa-large encoder with three independently resident RoBERTa-large encoders. Both configurations use FP32 model weights, BF16 autocast, batches of 32 CU--context pairs, and a 512-token pair length on an NVIDIA RTX PRO 6000 Blackwell. Encoder forward time is reported as the median of ten measurements following three warm-up passes.

As shown in Table~\ref{tab:resource-efficiency}, the shared configuration reduces the number of resident parameters from approximately 1.06 billion to 354 million, weight storage from 3.96 to 1.32~GiB, and peak allocated memory from 5.21 to 2.60~GiB. Per-batch encoder forward times remain nearly unchanged. The principal resource benefit therefore comes from maintaining one reusable encoder rather than three separate resident models, rather than from reducing the forward cost of an individual batch.

These measurements cover encoder residency and forward computation only. Graph construction, policy-guided hypernym mapping, candidate retrieval, model loading, and external model or API calls are not included.

\begin{table}[t]
\centering
\caption{Diagnostic of encoder residency and forward cost for one shared versus three independently resident RoBERTa-large encoders. Forward times use batches of 32 pairs with 512-token pair length and exclude graph construction, candidate retrieval, model loading, and external model or API calls.}
\label{tab:resource-efficiency}
\begin{tabular}{lrr}
\hline
Metric &
\begin{tabular}[c]{@{}c@{}}One shared\\model resident\end{tabular} &
\begin{tabular}[c]{@{}c@{}}Three independent\\models resident\end{tabular} \\
\hline
Parameters
& 354,313,219
& 1,062,933,507 \\
Weight storage (GiB)
& 1.3201
& 3.9602 \\
Peak allocated memory (GiB)
& 2.6024
& 5.2100 \\
EU2UK ms/batch
& 117.3967
& 117.2885 \\
GDPR-13 ms/batch
& 117.1851
& 117.0621 \\
CONTRACT ms/batch
& 117.2558
& 117.0978 \\
\hline
\end{tabular}
\end{table}

\section{Graph Fidelity Evaluation Details}
\label{app:fidelity}

\subsection{Cycle-Consistency Protocol}

RQ2 evaluates graph fidelity using reconstruction-based proxy measures because large-scale independently annotated Policy Graphs and Context Graphs are not available. Let \(T_0\) denote the source text and \(G_0\) the graph constructed from that text. A single reconstruction evaluates

\begin{equation}
T_0 \rightarrow G_0 \rightarrow T_1,
\end{equation}

where \(T_1\) is reconstructed from \(G_0\). The Cycle-Consistency Test extends this process by repeatedly applying

\begin{equation}
T_k \rightarrow G_k \rightarrow T_{k+1}.
\end{equation}

The objective is to assess whether the information represented in the graph remains stable as it repeatedly passes through text-to-graph and graph-to-text transformations.

Semantic preservation is measured using a symmetric maximum-similarity score over unit-normalized sentence embeddings. Let \(A\) and \(B\) denote the sentence-embedding sets obtained from the source and reconstructed texts, respectively. We compute

\begingroup
\fontsize{8pt}{11pt}\selectfont
\begin{equation}
\label{eq:cycle-semantic}
S(A,B)
=
\frac{1}{2}
\left[
\frac{1}{|A|}
\sum_{a\in A}
\max_{b\in B}
\cos(a,b)
+
\frac{1}{|B|}
\sum_{b\in B}
\max_{a\in A}
\cos(a,b)
\right].
\end{equation}
\endgroup

The symmetric formulation evaluates correspondence in both directions. It therefore reflects source information that is absent from the reconstruction as well as content that appears only in the reconstructed text.

Structural preservation is measured using the graph-level structural-consistency score used in the fidelity experiment. The score compares structural summary statistics between the initial and reconstructed graphs. It measures preservation of graph-level structure rather than exact node- and edge-level graph isomorphism.

\subsection{Noise-Injection Protocol}

Cycle consistency measures whether a graph representation remains stable under repeated reconstruction, but stability alone does not establish that the fidelity measure is sensitive to graph errors. A systematic omission, for example, may be reproduced consistently across multiple reconstruction cycles. We therefore complement cycle consistency with the Noise-Injection Test.

Starting from \(G_0\), we construct a corrupted graph \(G'_0\) by perturbing graph links and CU attributes. The perturbations include edge deletion, spurious-edge addition, and attribute corruption. Text \(T'_1\) is then reconstructed from \(G'_0\), and semantic similarity between \(T_0\) and \(T'_1\) is measured using the same symmetric embedding-based score defined in Eq.~\ref{eq:cycle-semantic}.

The experiment considers five corruption levels:

\begin{equation}
\delta \in \{0.00,\,0.03,\,0.05,\,0.10,\,0.20\}.
\end{equation}

As reported in RQ2, reconstruction similarity decreases monotonically as the corruption level increases. Together with the cycle-consistency results, this response provides complementary evidence that the graph representations are stable under repeated reconstruction while the fidelity measure remains sensitive to deliberate structural and attribute degradation.

\subsection{Scope of the Proxy Evaluation}

The Cycle-Consistency Test and Noise-Injection Test address complementary aspects of graph fidelity. Cycle consistency examines representation stability, whereas noise injection examines whether the same measurement responds when graph information is deliberately damaged. These tests are intended as proxy evaluations of information preservation rather than substitutes for direct comparison with independently annotated reference graphs.

A systematic extraction error can remain stable across reconstruction cycles, and high semantic similarity between source and reconstructed text does not imply exact logical or legal equivalence. Accordingly, the graph-fidelity results are interpreted as evidence of semantic and structural preservation under the evaluated procedure, rather than as direct certification of node- and edge-level graph correctness.
\clearpage
\bibliographystyle{IEEEtran}
\bibliography{references}

\begin{thebibliography}{10}
\providecommand{\url}[1]{#1}
\csname url@samestyle\endcsname
\providecommand{\newblock}{\relax}
\providecommand{\bibinfo}[2]{#2}
\providecommand{\BIBentrySTDinterwordspacing}{\spaceskip=0pt\relax}
\providecommand{\BIBentryALTinterwordstretchfactor}{4}
\providecommand{\BIBentryALTinterwordspacing}{\spaceskip=\fontdimen2\font plus
\BIBentryALTinterwordstretchfactor\fontdimen3\font minus
  \fontdimen4\font\relax}
\providecommand{\BIBforeignlanguage}[2]{{%
\expandafter\ifx\csname l@#1\endcsname\relax
\typeout{** WARNING: IEEEtran.bst: No hyphenation pattern has been}%
\typeout{** loaded for the language `#1'. Using the pattern for}%
\typeout{** the default language instead.}%
\else
\language=\csname l@#1\endcsname
\fi
#2}}
\providecommand{\BIBdecl}{\relax}
\BIBdecl

\bibitem{governatori2010norm}
\BIBentryALTinterwordspacing
G.~Governatori and A.~Rotolo, ``Norm compliance in business process modeling,''
  in \emph{Semantic Web Rules}, ser. Lecture Notes in Computer Science.\hskip
  1em plus 0.5em minus 0.4em\relax Springer, 2010, vol. 6403, pp. 194--209.
  [Online]. Available:
  \url{https://link.springer.com/chapter/10.1007/978-3-642-16289-3_17}
\BIBentrySTDinterwordspacing

\bibitem{athan2015legalruleml}
\BIBentryALTinterwordspacing
T.~Athan, G.~Governatori, M.~Palmirani, A.~Paschke, and A.~Wyner,
  ``{LegalRuleML}: Design principles and foundations,'' in \emph{Reasoning Web.
  Web Logic Rules}, ser. Lecture Notes in Computer Science.\hskip 1em plus
  0.5em minus 0.4em\relax Springer, 2015, vol. 9203, pp. 151--188. [Online].
  Available:
  \url{https://link.springer.com/chapter/10.1007/978-3-319-21768-0_6}
\BIBentrySTDinterwordspacing

\bibitem{saeidi2021crosspolicy}
\BIBentryALTinterwordspacing
M.~Saeidi, M.~Yazdani, and A.~Vlachos, ``Cross-policy compliance detection via
  question answering,'' in \emph{Proceedings of the 2021 Conference on
  Empirical Methods in Natural Language Processing}.\hskip 1em plus 0.5em minus
  0.4em\relax Association for Computational Linguistics, 2021, pp. 8622--8632.
  [Online]. Available: \url{https://aclanthology.org/2021.emnlp-main.678/}
\BIBentrySTDinterwordspacing

\bibitem{sun2025compliance}
\BIBentryALTinterwordspacing
J.~Sun, Z.~Luo, and Y.~Li, ``A compliance checking framework based on retrieval
  augmented generation,'' in \emph{Proceedings of the 31st International
  Conference on Computational Linguistics}.\hskip 1em plus 0.5em minus
  0.4em\relax Association for Computational Linguistics, 2025, pp. 2603--2615.
  [Online]. Available: \url{https://aclanthology.org/2025.coling-main.178/}
\BIBentrySTDinterwordspacing

\bibitem{echenim2025iot}
\BIBentryALTinterwordspacing
K.~U. Echenim and K.~P. Joshi, ``Automating {IoT} data privacy compliance by
  integrating knowledge graphs with large language models,'' \emph{IEEE
  Access}, vol.~13, pp. 118\,438--118\,451, 2025. [Online]. Available:
  \url{https://doi.org/10.1109/ACCESS.2025.3586278}
\BIBentrySTDinterwordspacing

\bibitem{lewis2020rag}
\BIBentryALTinterwordspacing
P.~Lewis, E.~Perez, A.~Piktus, F.~Petroni, V.~Karpukhin, N.~Goyal,
  H.~K{\"u}ttler, M.~Lewis, W.-t. Yih, T.~Rockt{\"a}schel, S.~Riedel, and
  D.~Kiela, ``Retrieval-augmented generation for knowledge-intensive {NLP}
  tasks,'' in \emph{Advances in Neural Information Processing Systems},
  vol.~33.\hskip 1em plus 0.5em minus 0.4em\relax Curran Associates, Inc.,
  2020, pp. 9459--9474. [Online]. Available:
  \url{https://proceedings.neurips.cc/paper/2020/hash/6b493230205f780e1bc26945df7481e5-Abstract.html}
\BIBentrySTDinterwordspacing

\bibitem{edge2024graphrag}
\BIBentryALTinterwordspacing
D.~Edge, H.~Trinh, N.~Cheng, J.~Bradley, A.~Chao, A.~Mody, S.~Truitt,
  D.~Metropolitansky, R.~O. Ness, and J.~Larson, ``From local to global: A
  graph {RAG} approach to query-focused summarization,'' \emph{arXiv preprint
  arXiv:2404.16130}, 2024. [Online]. Available:
  \url{https://arxiv.org/abs/2404.16130}
\BIBentrySTDinterwordspacing

\bibitem{koreeda2021contractnli}
\BIBentryALTinterwordspacing
Y.~Koreeda and C.~D. Manning, ``{ContractNLI}: A dataset for document-level
  natural language inference for contracts,'' in \emph{Findings of the
  Association for Computational Linguistics: EMNLP 2021}.\hskip 1em plus 0.5em
  minus 0.4em\relax Association for Computational Linguistics, Nov. 2021, pp.
  1907--1919. [Online]. Available:
  \url{https://aclanthology.org/2021.findings-emnlp.164/}
\BIBentrySTDinterwordspacing

\bibitem{gokhan2025rirag}
\BIBentryALTinterwordspacing
T.~Gokhan, K.~Wang, I.~Gurevych, and T.~Briscoe, ``Shared task {RIRAG}-2025:
  Regulatory information retrieval and answer generation,'' in
  \emph{Proceedings of the 1st Regulatory NLP Workshop (RegNLP 2025)}.\hskip
  1em plus 0.5em minus 0.4em\relax Association for Computational Linguistics,
  Jan. 2025, pp. 1--4. [Online]. Available:
  \url{https://aclanthology.org/2025.regnlp-1.1/}
\BIBentrySTDinterwordspacing

\bibitem{barati2020gdpr}
\BIBentryALTinterwordspacing
M.~Barati, O.~Rana, I.~Petri, and G.~Theodorakopoulos, ``{GDPR} compliance
  verification in internet of things,'' \emph{IEEE Access}, vol.~8, pp.
  119\,697--119\,709, 2020. [Online]. Available:
  \url{https://doi.org/10.1109/ACCESS.2020.3005509}
\BIBentrySTDinterwordspacing

\bibitem{guo2021semantic}
\BIBentryALTinterwordspacing
D.~Guo, E.~Onstein, and A.~D. La~Rosa, ``A semantic approach for automated rule
  compliance checking in construction industry,'' \emph{IEEE Access}, vol.~9,
  pp. 129\,648--129\,660, 2021. [Online]. Available:
  \url{https://doi.org/10.1109/ACCESS.2021.3108226}
\BIBentrySTDinterwordspacing

\bibitem{amaral2023dpa}
\BIBentryALTinterwordspacing
O.~A. Cejas, M.~I. Azeem, S.~Abualhaija, and L.~C. Briand, ``{NLP}-based
  automated compliance checking of data processing agreements against {GDPR},''
  \emph{IEEE Transactions on Software Engineering}, vol.~49, no.~9, pp.
  4282--4303, 2023. [Online]. Available:
  \url{https://doi.org/10.1109/TSE.2023.3288901}
\BIBentrySTDinterwordspacing

\bibitem{liu2021notified}
\BIBentryALTinterwordspacing
S.~Liu, B.~Zhao, R.~Guo, G.~Meng, F.~Zhang, and M.~Zhang, ``Have you been
  properly notified? automatic compliance analysis of privacy policy text with
  {GDPR} article 13,'' in \emph{Proceedings of the Web Conference 2021}.\hskip
  1em plus 0.5em minus 0.4em\relax Association for Computing Machinery, 2021,
  pp. 2154--2164. [Online]. Available:
  \url{https://impillar.github.io/files/www2021autocompliance.pdf}
\BIBentrySTDinterwordspacing

\bibitem{aires2018offsets}
\BIBentryALTinterwordspacing
J.~P. Aires, J.~Monteiro, R.~Granada, and F.~Meneguzzi, ``Norm conflict
  identification using vector space offsets,'' in \emph{2018 International
  Joint Conference on Neural Networks (IJCNN)}.\hskip 1em plus 0.5em minus
  0.4em\relax IEEE, 2018, pp. 1--8. [Online]. Available:
  \url{https://www.meneguzzi.eu/felipe/pubs/ijcnn-contract-conflicts-2018.pdf}
\BIBentrySTDinterwordspacing

\bibitem{aires2019contractual}
\BIBentryALTinterwordspacing
J.~P. Aires, R.~Granada, J.~Monteiro, R.~C. Barros, and F.~Meneguzzi,
  ``Classification of contractual conflicts via learning of semantic
  representations,'' in \emph{Proceedings of the 18th International Conference
  on Autonomous Agents and Multiagent Systems}.\hskip 1em plus 0.5em minus
  0.4em\relax IFAAMAS, 2019, pp. 1764--1766, extended abstract. [Online].
  Available: \url{https://www.ifaamas.org/Proceedings/aamas2019/pdfs/p1764.pdf}
\BIBentrySTDinterwordspacing

\bibitem{huang2024neuralconflict}
\BIBentryALTinterwordspacing
S.~Huang, J.~Sun, and R.~Li, ``{NeuralConflict}: Using neural networks to
  identify norm conflicts in normative documents,'' \emph{Expert Systems},
  vol.~41, no.~6, p. e13035, 2024. [Online]. Available:
  \url{https://onlinelibrary.wiley.com/doi/10.1111/exsy.13035}
\BIBentrySTDinterwordspacing

\bibitem{gutierrez2024hipporag}
\BIBentryALTinterwordspacing
B.~J. Guti{\'e}rrez, Y.~Shu, Y.~Gu, M.~Yasunaga, and Y.~Su, ``{HippoRAG}:
  Neurobiologically inspired long-term memory for large language models,'' in
  \emph{Advances in Neural Information Processing Systems}, vol.~37.\hskip 1em
  plus 0.5em minus 0.4em\relax Curran Associates, Inc., 2024, pp.
  59\,532--59\,569. [Online]. Available:
  \url{https://proceedings.neurips.cc/paper_files/paper/2024/hash/6ddc001d07ca4f319af96a3024f6dbd1-Abstract-Conference.html}
\BIBentrySTDinterwordspacing

\bibitem{he2024gretriever}
\BIBentryALTinterwordspacing
X.~He, Y.~Tian, Y.~Sun, N.~V. Chawla, T.~Laurent, Y.~LeCun, X.~Bresson, and
  B.~Hooi, ``{G-Retriever}: Retrieval-augmented generation for textual graph
  understanding and question answering,'' in \emph{Advances in Neural
  Information Processing Systems}, vol.~37.\hskip 1em plus 0.5em minus
  0.4em\relax Curran Associates, Inc., 2024. [Online]. Available:
  \url{https://proceedings.neurips.cc/paper_files/paper/2024/hash/efaf1c9726648c8ba363a5c927440529-Abstract-Conference.html}
\BIBentrySTDinterwordspacing

\bibitem{elluri2021bert}
\BIBentryALTinterwordspacing
L.~Elluri, S.~S.~L. Chukkapalli, K.~P. Joshi, T.~Finin, and A.~Joshi, ``A
  {BERT} based approach to measure web services policies compliance with
  {GDPR},'' \emph{IEEE Access}, vol.~9, pp. 148\,004--148\,016, 2021. [Online].
  Available: \url{https://doi.org/10.1109/ACCESS.2021.3123950}
\BIBentrySTDinterwordspacing

\bibitem{garza2024privcompkg}
\BIBentryALTinterwordspacing
L.~Garza, L.~Elluri, A.~Kotal, A.~Piplai, D.~Gupta, and A.~Joshi,
  ``{PrivComp-KG}: Leveraging knowledge graph and large language models for
  privacy policy compliance verification,'' \emph{arXiv preprint
  arXiv:2404.19744}, 2024. [Online]. Available:
  \url{https://arxiv.org/abs/2404.19744}
\BIBentrySTDinterwordspacing

\bibitem{cui2023poligraph}
\BIBentryALTinterwordspacing
H.~Cui, R.~Trimananda, A.~Markopoulou, and S.~Jordan, ``{PoliGraph}: Automated
  privacy policy analysis using knowledge graphs,'' in \emph{32nd USENIX
  Security Symposium (USENIX Security 23)}.\hskip 1em plus 0.5em minus
  0.4em\relax USENIX Association, Aug. 2023, pp. 1037--1054. [Online].
  Available:
  \url{https://www.usenix.org/conference/usenixsecurity23/presentation/cui}
\BIBentrySTDinterwordspacing

\bibitem{chalkidis2021doc2doc}
\BIBentryALTinterwordspacing
I.~Chalkidis, M.~Fergadiotis, N.~Manginas, E.~Katakalou, and P.~Malakasiotis,
  ``Regulatory compliance through {Doc2Doc} information retrieval: A case study
  in {EU/UK} legislation where text similarity has limitations,'' in
  \emph{Proceedings of the 16th Conference of the European Chapter of the
  Association for Computational Linguistics: Main Volume}.\hskip 1em plus 0.5em
  minus 0.4em\relax Association for Computational Linguistics, 2021, pp.
  3498--3511. [Online]. Available:
  \url{https://aclanthology.org/2021.eacl-main.305/}
\BIBentrySTDinterwordspacing

\bibitem{zhang2023bertgdpr}
\BIBentryALTinterwordspacing
L.~Zhang, N.~Moukafih, H.~Alamri, G.~Epiphaniou, and C.~Maple, ``A {BERT}-based
  empirical study of privacy policies' compliance with {GDPR},'' in \emph{2023
  IEEE Conference on Communications and Network Security (CNS)}.\hskip 1em plus
  0.5em minus 0.4em\relax IEEE, 2023, pp. 1--6. [Online]. Available:
  \url{https://ieeexplore.ieee.org/document/10288797/}
\BIBentrySTDinterwordspacing

\bibitem{ichida2021contract}
\BIBentryALTinterwordspacing
A.~Y. Ichida and F.~Meneguzzi, ``Detecting logical relation in contract
  clauses,'' \emph{arXiv preprint arXiv:2111.01856}, 2021. [Online]. Available:
  \url{https://arxiv.org/abs/2111.01856}
\BIBentrySTDinterwordspacing

\bibitem{sheng2026dynamic}
\BIBentryALTinterwordspacing
W.~Sheng, C.~Wei, Y.~Zhang, B.~Zhang, and J.~Sun, ``A dynamic prompt-based
  logic-aided compliance checker,'' \emph{Big Data and Cognitive Computing},
  vol.~10, no.~3, p.~95, 2026. [Online]. Available:
  \url{https://www.mdpi.com/2504-2289/10/3/95}
\BIBentrySTDinterwordspacing

\bibitem{openai2025gpt41}
\BIBentryALTinterwordspacing
{OpenAI}, ``Introducing {GPT-4.1} in the {API},'' Apr. 2025, accessed: Sep. 30,
  2026. [Online]. Available: \url{https://openai.com/index/gpt-4-1/}
\BIBentrySTDinterwordspacing

\bibitem{mo2025kggen}
\BIBentryALTinterwordspacing
B.~Mo, K.~Yu, J.~Kazdan, J.~Cabezas, P.~Mpala, L.~Yu, C.~Cundy, C.~Kanatsoulis,
  and S.~Koyejo, ``{KGGen}: Extracting knowledge graphs from plain text with
  language models,'' in \emph{Advances in Neural Information Processing
  Systems}, vol.~38.\hskip 1em plus 0.5em minus 0.4em\relax Curran Associates,
  Inc., 2025. [Online]. Available:
  \url{https://proceedings.neurips.cc/paper_files/paper/2025/file/2b368455e832d2b1a60bcad8c4c6481f-Paper-Conference.pdf}
\BIBentrySTDinterwordspacing

\bibitem{yang2025qwen3}
\BIBentryALTinterwordspacing
A.~Yang, A.~Li, B.~Yang, B.~Zhang, B.~Hui, B.~Zheng \emph{et~al.}, ``{Qwen3}
  technical report,'' \emph{arXiv preprint arXiv:2505.09388}, 2025. [Online].
  Available: \url{https://arxiv.org/abs/2505.09388}
\BIBentrySTDinterwordspacing

\bibitem{gdpr2016}
\BIBentryALTinterwordspacing
{European Parliament and Council of the European Union}, ``Regulation ({EU})
  2016/679 of the european parliament and of the council of 27 april 2016 on
  the protection of natural persons with regard to the processing of personal
  data and on the free movement of such data, and repealing directive
  95/46/{EC} ({General Data Protection Regulation}),'' Official Journal of the
  European Union, L 119, pp. 1--88, May 2016. [Online]. Available:
  \url{https://eur-lex.europa.eu/eli/reg/2016/679/oj}
\BIBentrySTDinterwordspacing

\bibitem{dietterich1997mil}
\BIBentryALTinterwordspacing
T.~G. Dietterich, R.~H. Lathrop, and T.~Lozano-P{\'e}rez, ``Solving the
  multiple instance problem with axis-parallel rectangles,'' \emph{Artificial
  Intelligence}, vol.~89, no. 1--2, pp. 31--71, 1997. [Online]. Available:
  \url{https://www.sciencedirect.com/science/article/pii/S0004370296000343}
\BIBentrySTDinterwordspacing

\bibitem{ilse2018mil}
\BIBentryALTinterwordspacing
M.~Ilse, J.~Tomczak, and M.~Welling, ``Attention-based deep multiple instance
  learning,'' in \emph{Proceedings of the 35th International Conference on
  Machine Learning}, ser. Proceedings of Machine Learning Research,
  vol.~80.\hskip 1em plus 0.5em minus 0.4em\relax PMLR, 2018, pp. 2127--2136.
  [Online]. Available: \url{https://proceedings.mlr.press/v80/ilse18a.html}
\BIBentrySTDinterwordspacing

\bibitem{wehnert2022bert}
\BIBentryALTinterwordspacing
S.~Wehnert, S.~Dureja, L.~Kutty, V.~Sudhi, and E.~W. De~Luca, ``Applying {BERT}
  embeddings to predict legal textual entailment,'' \emph{The Review of
  Socionetwork Strategies}, vol.~16, no.~1, pp. 197--219, 2022. [Online].
  Available: \url{https://link.springer.com/article/10.1007/s12626-022-00101-3}
\BIBentrySTDinterwordspacing

\bibitem{sun2017inner}
\BIBentryALTinterwordspacing
C.~Sun, Y.~Liu, C.~Jia, B.~Liu, and L.~Lin, ``Recognizing text entailment via
  bidirectional {LSTM} model with inner-attention,'' in \emph{Intelligent
  Computing Methodologies}.\hskip 1em plus 0.5em minus 0.4em\relax Springer,
  2017, pp. 448--457. [Online]. Available:
  \url{https://link.springer.com/chapter/10.1007/978-3-319-63315-2_39}
\BIBentrySTDinterwordspacing

\bibitem{chen2019knowledge}
\BIBentryALTinterwordspacing
D.~Chen, Y.~Li, M.~Yang, H.-T. Zheng, and Y.~Shen, ``Knowledge-aware textual
  entailment with graph attention network,'' in \emph{Proceedings of the 28th
  ACM International Conference on Information and Knowledge Management}.\hskip
  1em plus 0.5em minus 0.4em\relax Association for Computing Machinery, 2019,
  pp. 2145--2148. [Online]. Available:
  \url{https://dl.acm.org/doi/10.1145/3357384.3358071}
\BIBentrySTDinterwordspacing

\bibitem{openai2025gpt5systemcard}
\BIBentryALTinterwordspacing
{OpenAI}, ``{GPT-5} system card,'' Aug. 2025, accessed: Sep. 30, 2026.
  [Online]. Available: \url{https://openai.com/index/gpt-5-system-card/}
\BIBentrySTDinterwordspacing

\bibitem{liu2019roberta}
\BIBentryALTinterwordspacing
Y.~Liu, M.~Ott, N.~Goyal, J.~Du, M.~Joshi, D.~Chen, O.~Levy, M.~Lewis,
  L.~Zettlemoyer, and V.~Stoyanov, ``{RoBERTa}: A robustly optimized {BERT}
  pretraining approach,'' \emph{arXiv preprint arXiv:1907.11692}, 2019.
  [Online]. Available: \url{https://arxiv.org/abs/1907.11692}
\BIBentrySTDinterwordspacing

\bibitem{chen2024m3}
\BIBentryALTinterwordspacing
J.~Chen, S.~Xiao, P.~Zhang, K.~Luo, D.~Lian, and Z.~Liu, ``{M3}-embedding:
  Multi-linguality, multi-functionality, multi-granularity text embeddings
  through self-knowledge distillation,'' in \emph{Findings of the Association
  for Computational Linguistics: ACL 2024}.\hskip 1em plus 0.5em minus
  0.4em\relax Association for Computational Linguistics, Aug. 2024, pp.
  2318--2335. [Online]. Available:
  \url{https://aclanthology.org/2024.findings-acl.137/}
\BIBentrySTDinterwordspacing

\bibitem{baai_reranker_v2_m3}
\BIBentryALTinterwordspacing
{BAAI}, ``{BGE-Reranker-v2-M3}: Model card,'' Hugging Face, accessed: Sep. 30,
  2026. [Online]. Available:
  \url{https://huggingface.co/BAAI/bge-reranker-v2-m3}
\BIBentrySTDinterwordspacing

\bibitem{loshchilov2019adamw}
\BIBentryALTinterwordspacing
I.~Loshchilov and F.~Hutter, ``Decoupled weight decay regularization,'' in
  \emph{International Conference on Learning Representations}, 2019. [Online].
  Available: \url{https://arxiv.org/abs/1711.05101}
\BIBentrySTDinterwordspacing

\bibitem{chicco2020mcc}
\BIBentryALTinterwordspacing
D.~Chicco and G.~Jurman, ``The advantages of the {Matthews} correlation
  coefficient ({MCC}) over {F1} score and accuracy in binary classification
  evaluation,'' \emph{BMC Genomics}, vol.~21, p.~6, 2020. [Online]. Available:
  \url{https://bmcgenomics.biomedcentral.com/articles/10.1186/s12864-019-6413-7}
\BIBentrySTDinterwordspacing

\bibitem{he2021debertav3}
\BIBentryALTinterwordspacing
P.~He, J.~Gao, and W.~Chen, ``{DeBERTaV3}: Improving {DeBERTa} using
  {ELECTRA}-style pre-training with gradient-disentangled embedding sharing,''
  \emph{arXiv preprint arXiv:2111.09543}, 2021. [Online]. Available:
  \url{https://arxiv.org/abs/2111.09543}
\BIBentrySTDinterwordspacing

\bibitem{warner2025modernbert}
\BIBentryALTinterwordspacing
B.~Warner, A.~Chaffin, B.~Clavi{\'e}, O.~Weller, O.~Hallstr{\"o}m,
  S.~Taghadouini, A.~Gallagher, R.~Biswas, F.~Ladhak, T.~Aarsen, G.~T. Adams,
  J.~Howard, and I.~Poli, ``Smarter, better, faster, longer: A modern
  bidirectional encoder for fast, memory efficient, and long context finetuning
  and inference,'' in \emph{Proceedings of the 63rd Annual Meeting of the
  Association for Computational Linguistics (Volume 1: Long Papers)}.\hskip 1em
  plus 0.5em minus 0.4em\relax Association for Computational Linguistics, 2025,
  pp. 2526--2547. [Online]. Available:
  \url{https://aclanthology.org/2025.acl-long.127/}
\BIBentrySTDinterwordspacing

\bibitem{chalkidis2020legalbert}
\BIBentryALTinterwordspacing
I.~Chalkidis, M.~Fergadiotis, P.~Malakasiotis, N.~Aletras, and
  I.~Androutsopoulos, ``{LEGAL-BERT}: The muppets straight out of law school,''
  in \emph{Findings of the Association for Computational Linguistics: EMNLP
  2020}.\hskip 1em plus 0.5em minus 0.4em\relax Association for Computational
  Linguistics, 2020, pp. 2898--2904. [Online]. Available:
  \url{https://aclanthology.org/2020.findings-emnlp.261/}
\BIBentrySTDinterwordspacing

\end{thebibliography}

\makeatletter
\renewcommand{\@IEEEBIOskipN}{12pt}

\xpatchcmd{\IEEEbiography}
  {\vskip \@IEEEBIOskipN plus 1fil minus 0\baselineskip}
  {\vskip \@IEEEBIOskipN}
  {}
  {\PackageError{BiographySpacing}
    {Photo biography spacing patch failed}
    {Check the IEEEbiography definition}}
\makeatother

\raggedbottom

\patchcmd{\IEEEbiographynophoto}
  {\vskip 4\baselineskip plus 1fil minus 0\baselineskip}
  {\vskip 12pt}
  {}
  {\PackageError{GraphCompliance}{Could not set biography spacing}{Inspect the IEEE biography environment}}
\raggedbottom
\patchcmd{\IEEEbiographynophoto}
  {\interlinepenalty500}
  {\interlinepenalty10000}
  {}
  {\PackageError{GraphCompliance}{Could not keep biography paragraphs together}{Inspect the IEEE biography environment}}
\section*{Author biographies}

\begin{IEEEbiography}[{
  \includegraphics[
    width=1in,
    height=1.25in,
    keepaspectratio
  ]{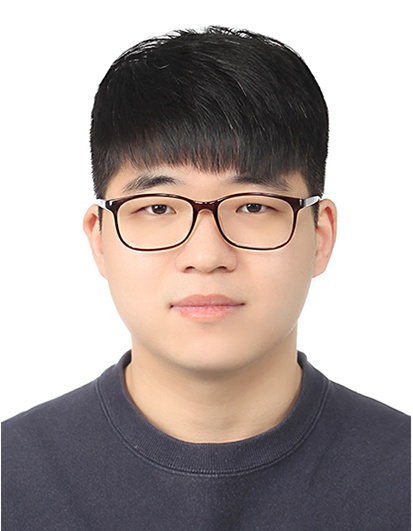}
}]{Jiseong Chung}
is currently pursuing the Ph.D. degree with the Department of Naval Architecture and Ocean Engineering, Seoul National University, Seoul, Republic of Korea. He is a member of the Deep Learning Laboratory at Seoul National University. His research interests include retrieval-augmented generation, edge-optimized language models, GraphRAG, and AI compliance, with broader interests in efficient and trustworthy artificial intelligence systems.
\end{IEEEbiography}

\begin{IEEEbiography}[{
  \includegraphics[
    width=1in,
    height=1.25in,
    keepaspectratio
  ]{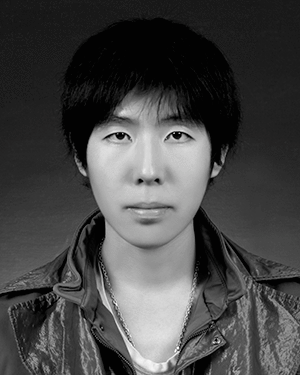}
}]{Ronny Ko}
received the Ph.D. degree in computer science from Harvard University, Cambridge, MA, USA, in 2021. He is currently a Specially Appointed Assistant Professor with the Graduate School of Information Science and Technology, The University of Osaka, Osaka, Japan. His research interests include database simulation, what-if analysis, efficient database replay, systems security and privacy, and multimodal Web agents.
\end{IEEEbiography}

\begin{IEEEbiography}[{
  \includegraphics[
    width=1in,
    height=1.25in,
    keepaspectratio
  ]{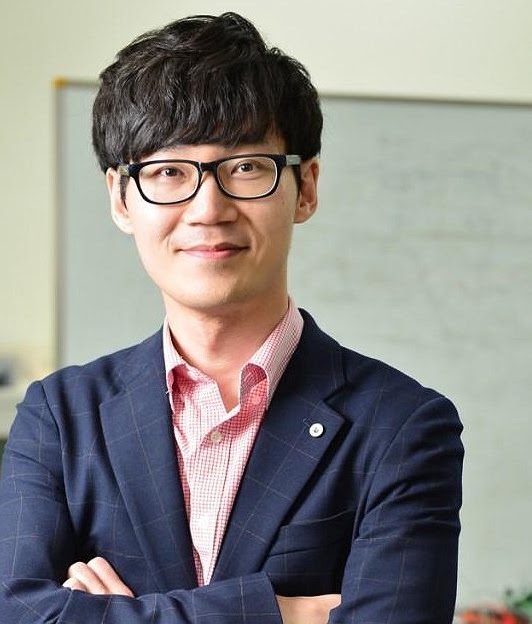}
}]{Won-Yong Shin}
received the Ph.D. degree in electrical engineering and computer science from the Korea Advanced Institute of Science and Technology (KAIST), Daejeon, Republic of Korea, in 2008. He was a Postdoctoral Fellow and subsequently a Research Associate with the School of Engineering and Applied Sciences, Harvard University, Cambridge, MA, USA. From 2012 to 2019, he was a faculty member with Dankook University, Yongin, Republic of Korea. Since 2019, he has been with the School of Mathematics and Computing (Computational Science and Engineering), Yonsei University, Seoul, Republic of Korea, where he is currently a Professor. His research interests include data mining, machine learning, graph and social-network analysis, information retrieval, learning-based design, healthcare informatics, bioinformatics, and wireless networking.
\end{IEEEbiography}

\begin{IEEEbiography}[{
  \includegraphics[
    width=1in,
    height=1.25in,
    keepaspectratio
  ]{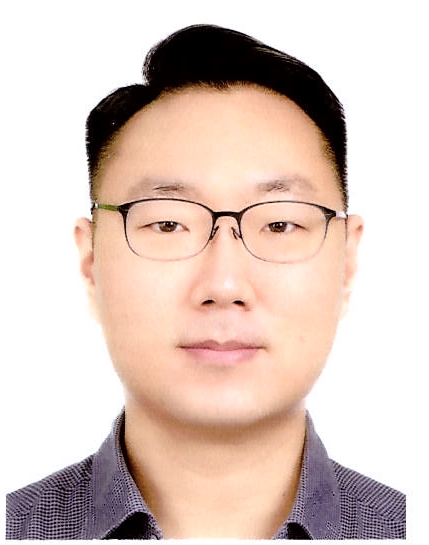}
}]{Wonchul Yoo}
received the Ph.D. degree in engineering from Seoul National University, Seoul, Republic of Korea, in 2022. His doctoral research focused on nautical route clustering using statistical trajectory-distance metrics. His research interests include route planning, machine learning, three-dimensional point-cloud processing, and data-driven computational methods for maritime navigation.
\end{IEEEbiography}

\begin{IEEEbiography}[{
  \includegraphics[
    width=1in,
    height=1.25in,
    keepaspectratio
  ]{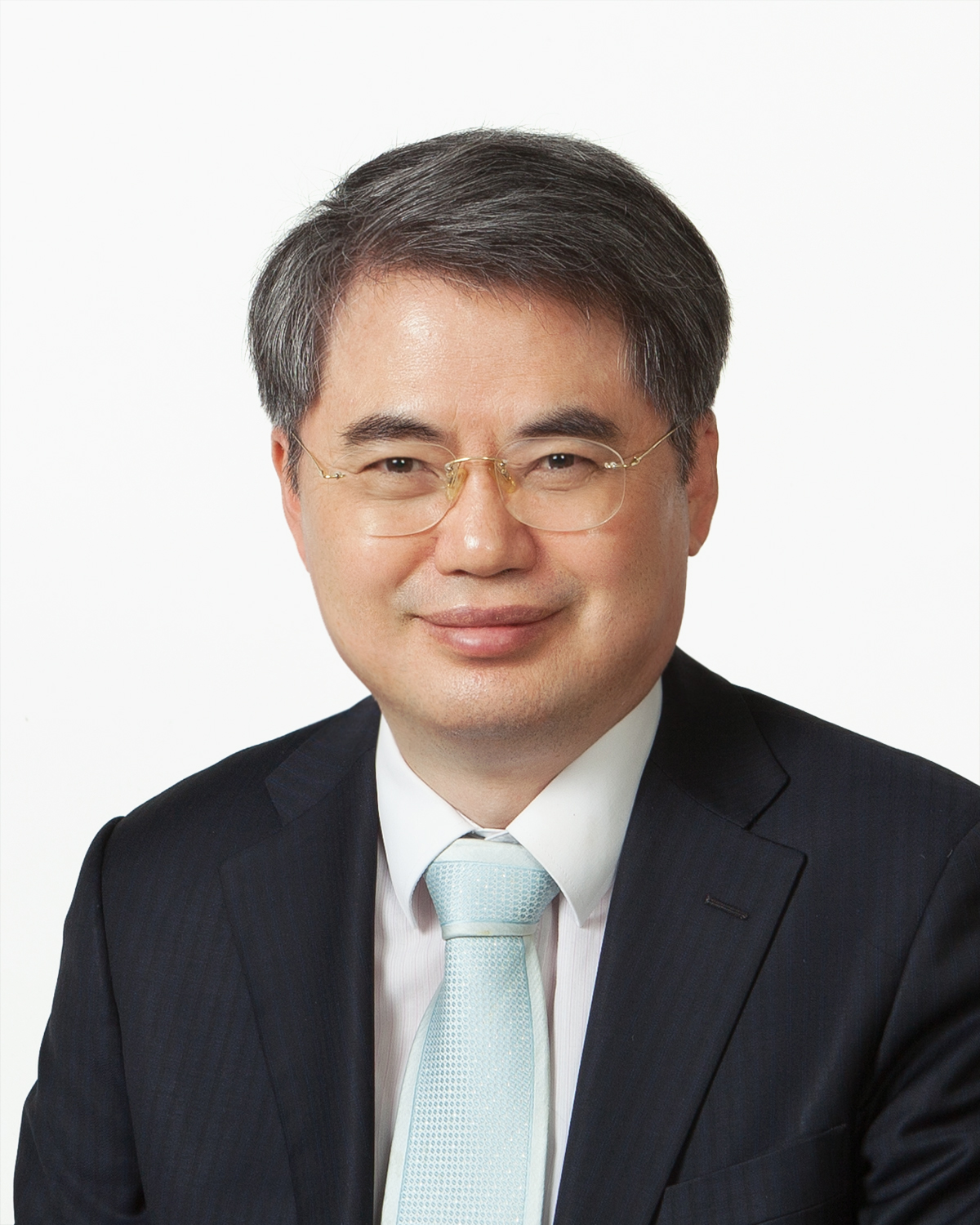}
}]{Tae-wan Kim}
received the Ph.D. degree in computer science from Arizona State University, Tempe, AZ, USA, in 1996. He subsequently worked with Structural Dynamics Research Corporation (SDRC), now part of Siemens Digital Industries Software, developing geometric-modeling technologies for commercial CAD/CAM/CAE systems. Since 2003, he has been a Professor with the Department of Naval Architecture and Ocean Engineering, Seoul National University, Seoul, Republic of Korea. He was a Visiting Scholar with the Massachusetts Institute of Technology, Cambridge, MA, USA, from 2018 to 2019. His research interests include large language models, privacy-preserving AI, deep learning, machine learning, reinforcement learning, simultaneous localization and mapping, and computer-aided design.
\end{IEEEbiography}

\EOD
\end{document}